\documentclass{interact}
\usepackage{tabularx}
\usepackage[utf8]{inputenc}
\usepackage[caption=false]{subfig}
\usepackage{graphicx}%
\usepackage{multirow}%
\usepackage{amsmath,amssymb,amsfonts}%
\usepackage{mathrsfs}%
\usepackage[title]{appendix}%
\usepackage{xcolor}%
\usepackage{textcomp}%
\usepackage{manyfoot}%
\usepackage{booktabs}%
\usepackage{algorithm}%
\usepackage{algorithmicx}%
\usepackage{algpseudocode}%
\usepackage{listings}%
\usepackage[section]{placeins}
\usepackage{tabularx}
\usepackage{rotating}
\usepackage{array}
\usepackage{hyperref}
\usepackage{comment}
\usepackage{textcomp}
\usepackage{longtable}
\hypersetup{
     colorlinks=true,
     linkcolor=red,
     filecolor=blue,
     citecolor = green!70!black,
     urlcolor=cyan,
     }

\usepackage[numbers,sort&compress]{natbib}
\bibpunct[, ]{[}{]}{,}{n}{,}{,}

\theoremstyle{plain}
\newtheorem{theorem}{Theorem}[section]
\newtheorem{lemma}[theorem]{Lemma}

\newtheorem*{assumption*}{Assumption}

\theoremstyle{definition}

\newcommand{\R}{\mathbb{R}}
\theoremstyle{remark}
\newtheorem{remark}{Remark}

\newcommand{\taux}{$\tau(X)$ }

\newcommand{\ie}{i.e., }
\newcommand{\eg}{e.g., }
\newcommand{\numtrees}{\textit{num.trees }}
\newcommand{\minnodesize}{\textit{min.node.size }}
\newcommand{\mtry}{\textit{mtry }}
\def\argmin{\mathop{\arg\min}}

\begin{document}

\articletype{Research Article}

\title{
A New Causal Rule Learning Approach to Interpretable Estimation of Heterogeneous Treatment Effect
}

\author{
\name{Ying Wu\textsuperscript{a}\thanks{CONTACT Ying Wu: yingwu@stu.xjtu.edu.cn; Hanzhong Liu: lhz2016@tsinghua.edu.cn; Kai Ren: rk90108@163.com; Shujie Ma: shujie.ma@ucr.edu; Xiangyu Chang: xiangyuchang@xjtu.edu.cn},
Hanzhong Liu\textsuperscript{b}, Kai Ren\textsuperscript{c}, Shujie Ma\textsuperscript{d} and Xiangyu Chang\textsuperscript{a}}
\affil{
\textsuperscript{a}Department of Information Systems and Intelligent Business, School of Management, Xi'an Jiaotong University, Xi’an, 710049, Shaanxi, P.R.China; \textsuperscript{b} Department of Industrial Engineering, Tsinghua University, 100084, Beijing, P.R.China;
\textsuperscript{c} Department of Cardiovascular Surgery, Xijing Hospital, Air Force Military Medical University, Xi’an, 710032, Shaanxi, P.R.China;
\textsuperscript{d} Department of Statistics, University of California-Riverside, Riverside, CA 92521, U.S.A.
}
}

\maketitle

\begin{abstract}
Interpretability plays a crucial role in the application of statistical learning to estimate heterogeneous treatment effects (HTE) in complex diseases. In this study, we leverage a rule-based workflow, namely causal rule learning (CRL), to estimate and improve our understanding of HTE for atrial septal defect, addressing an overlooked question in the previous literature: what if an individual simultaneously belongs to multiple groups with different average treatment effects? The CRL process consists of three steps: rule discovery, which generates a set of causal rules with corresponding subgroup average treatment effects; rule selection, which identifies a subset of these rules to deconstruct individual-level treatment effects as a linear combination of subgroup-level effects; and rule analysis, which presents a detailed procedure for further analyzing each selected rule from multiple perspectives to identify the most promising rules for validation.
Extensive simulation studies and real-world data analysis demonstrate that CRL outperforms other methods in providing interpretable estimates of HTE, especially when dealing with complex ground truth and sufficient sample sizes.
\end{abstract}


\section{Introduction}
\label{Intro}

In clinical research and practice, patients respond differently to specific treatment interventions, necessitating the exploration of heterogeneous treatment effects (HTE) to inform personalized treatment plans and optimize health outcomes.
From a practitioner's perspective, this exploration requires not only accurate estimation but also a deep understanding of HTE, which can be challenging to achieve in real-world clinical settings \cite{crab2022benchmarking}, especially for complex diseases such as atrial septal defect (ASD).
As a common type of congenital heart disease, ASD exhibits diverse pathologies and patient heterogeneity \cite{brida2021atrial}.
With an accurate estimation of HTE and improved interpretation of HTE estimates, cardiologists can better understand, trust, and justify their treatment recommendations, thereby encouraging effective doctor-patient communication and collaboration \cite{herlitz2017comparativism, crab2022benchmarking}.

So far, researchers have made great efforts to estimate HTE accurately using various methods, among which statistical learning methods are particularly popular due to their ability to handle high-dimensional datasets and uncover non-linear relationships between covariates \cite{wager2018estimation,zhao2012estimating, seibold2016model,zhou2017residual}. 
However, most methods are typically designed for general estimation purposes, lacking specific adaptations for clinical applications.
Besides, they seldom provide guidance on how to interpret what their model has learned from the data.
Consequently, clinicians may struggle to understand these methods, let alone apply them in real-world practice.

\begin{figure}[t]
	\centering
	\includegraphics[scale = 0.1]{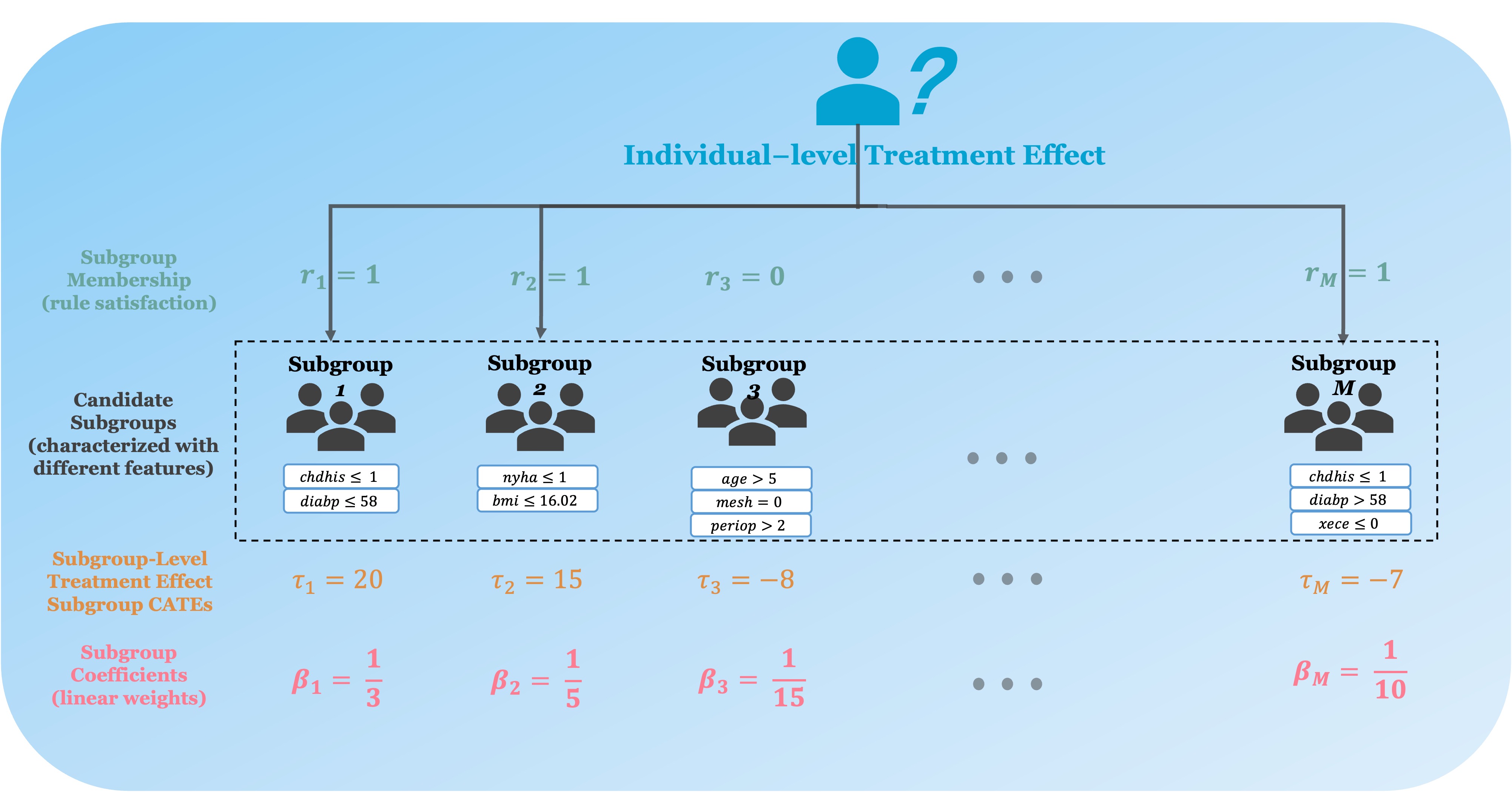}	
	\caption{Connection between individual-level and subgroup-level treatment effects.}
	\label{fig:motivation}
\end{figure}

There has been an increasing amount of research addressing the interpretability issue in HTE estimation \cite{dwivedi2020stable,hapfelmeier2018subgroup,subgrp2017huang, wang2021causal,chen2017general,foster2011subgroup}. The majority of this research focuses on identifying subgroups with enhanced (positive) treatment effects.
However, the setup and methodology of these studies may be far from sufficient to reveal the complex nature of certain diseases.
For complex disease treatment whose outcome is influenced by a lot of factors and interactions among these factors, we may find the situation shown in Figure \ref{fig:motivation}:
There are $M$ true subgroups (from subgroup 1 to subgroup $M$) that contribute to treatment effect heterogeneity.
The population within each subgroup (\eg, subgroup 1) shares the same treatment effect ($\tau_1 = 20$), \ie the subgroup's conditional average treatment effect (CATE) derived from the difference between the two treatment groups within the subgroup.
The subgroup CATEs between groups vary in effect size (or magnitude) and sign ($\tau_2 = 15$ and $\tau_3 = -8$).
Besides, these subgroups are characterized with (not necessarily) different features, such as subgroup 1 by $chdhis \le 1$ and $diabp \le 58$ or subgroup 2 by $NYHA \le 1$ and $BMI \le 16.02$.
Therefore, when we find the individual belongs to subgroups 1, 2, and $M$ ($r_1 = 1, r_2 = 1, r_M = 1$) at the same time (multiple memberships), it raises the question of what the individual-level treatment effect (ITE)
should be, given that the three subgroups differ in treatment effect. 
Considering the complex nature of certain diseases, we assume ITE as \textit{a linear combination of the above candidate group CATE} in this study.
Hence, the ITE in Figure \ref{fig:motivation} is $1/3 \times 20 + 1/5 \times 15 + 1/10 \times (-7) = 173/21$ where $\beta_1 = 1/3$$,$ $\beta_2 = 1/5$, and $\beta_M = 1/10$ are the linear coefficients in the combination which quantifies the contribution of each candidate subgroup to ITE.

Compared with our perspective, the studies focus on subgroup identification usually end up with a limited number of subgroups for interpretability concerns \cite{chen2023estimating,vegetabile2021distinction} (e.g., only subgroups 1 and 2 in Figure \ref{fig:motivation} are identified).
Some approaches even define mutually disjoint subgroups (e.g., \cite{wang2021causal}): One individual can only belong to one subgroup. 
As indicated in \cite{vegetabile2021distinction}, individual effects are not guaranteed to have even the same sign as CATE, let alone the same magnitude.
Therefore, an ITE cannot be determined solely by belonging to a few specific groups, which relates to a few covariates.
On the contrary, ITE may involve a broader range of subgroup memberships, and we need to consider how being in these subgroups collectively influences an individual, \ie to explore the relationship between individual-level treatment effects and group-level treatment effects.
In addition, since these studies emphasize a few subgroups, they are likely to ignore the estimation and understanding of ITE for the population outside the identified subgroups.

\begin{figure}[h]
	\centering
	\includegraphics[scale = 0.13]{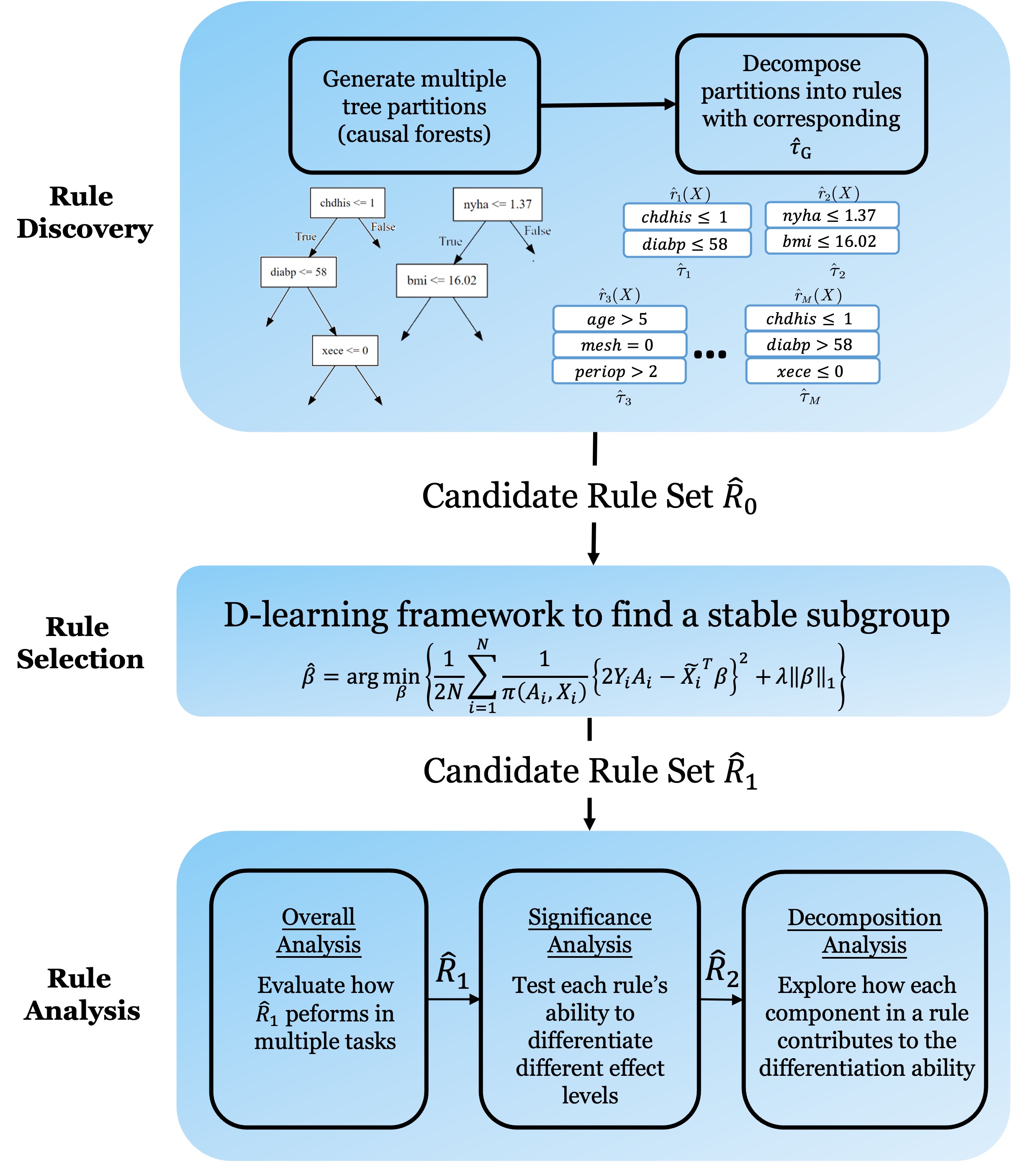}	
	\caption{Workflow of causal rule learning.}
	\label{fig:workflow}
\end{figure}

With consideration of the above, this study introduces a novel method, Causal Rule Learning (CRL), to integrate both accurate estimation and an in-depth understanding of HTE for complex diseases.
Guided by the Predictive, Descriptive, Relevant (PDR) framework of interpretability \cite{yubinPDR2019} (see Appendix \ref{apx:PDR} for more information) and inspired by the RuleFit \cite{friedman2008predictive} methodology, CRL caters to healthcare practitioners via an easily understood rule-based approach and an enlightening perspective on the relationship between ITE and CATE.
Specifically, CRL incorporates three steps, \ie rule discovery, rule selection, and rule analysis. 
The rule discovery step first generates a set of causal rules (characterized by different patient covariates) and their corresponding subgroup CATE estimates using causal forest \cite{wager2018estimation}. 
Then, the rule selection step leverages a D-learning method \cite{qi2018d} to filter out non-informative rules and use the informative ones to estimate ITE.
Lastly, the rule analysis step evaluates the rules from three different granularity levels, namely overall analysis, significance analysis, and decomposition analysis.
The whole workflow with details is shown in Figure \ref{fig:workflow}.

CRL contributes to existing literature and healthcare practice by enriching the toolset for interpretable HTE estimation, particularly in scenarios involving complex disease treatments.
In detail, we propose:
\begin{enumerate}
    \item A comprehensively interpretable framework for HTE estimation. CRL presents a rule-based workflow that includes the discovery, selection, and analysis of rules. 
Rules are used for their ease in human understanding,  usefulness in healthcare practice, and inclusion of potential knowledge.
The discovery step seeks a large number of rules, which are meaningful for revealing the unknown aspects of HTE.
The following selection step imposes sparsity on these rules, surfacing the most predictive and promising ones for ITE and further analysis.
The analysis step provides an in-depth investigation of the rules identified, ranging from the overall performance of the rules as a whole on HTE-related tasks to how each component contributes to a specific rule in differentiating effect magnitudes, providing an in-depth interpretation of the subgroups identified.

    \item A novel perspective integrating group and individual-level treatment effects: ITE as a linear combination of subgroup CATE.
This perspective attempts to explore and make clear the relationship between ITE and CATE, which answers a previously overlooked yet practically relevant question: What occurs when an individual patient belongs to multiple subgroups with different average treatment effects? 
By bridging subgroup characteristics with individual-level outcomes, we can build a deeper and more structured understanding of treatment heterogeneity that may align with the truth of complex diseases.

\end{enumerate}

Together, the refined rules, their corresponding subgroup CATEs, and weights (coefficients in the linear combination) obtained after the whole workflow 
provide insights to HTE estimation and benefit relevant clinical practice, such as inspiring the stratification design of related RCTs and clinical decision support systems, as illustrated later in our real-world application of CRL.

The rest of this paper is organized as follows: In Section \ref{related work}, we briefly summarize previous studies on HTE for clinical purposes. 
Section \ref{method} delineates the proposed CRL workflow, including its basic data settings and assumptions, formulation, and workflow. 
Section \ref{evaluation} demonstrates the effectiveness of CRL on both simulated data and real-world data. 
Section \ref{conclusion} concludes our work and discusses its limitations and potential future research directions.
For brevity in this paper, we will slightly misuse the following terms: i) HTE, CATE, and ITE; ii) \textit{satisfy a rule} and \textit{belongs to a subgroup}. These terms are used according to specific contexts and sometimes indiscriminately in broader situations.
\section{Related work}\label{related work}
This section presents extant literature related to HTE estimation from various perspectives.
The majority of studies focus on the estimation of CATE, namely the average treatment effect on specific subgroups.
A small stream of research focuses on treatment recommendation, \ie directly modeling the best individualized treatment rules or policy given personal characteristics to maximize the outcome of interest. 
Weighting methods are frequently used for this goal, such as \cite{zhao2012estimating, HE2021107167, zhou2017residual,li2024penalized}. 
Though the latter is capable of providing actionable suggestions to clinical treatment decisions, they often fail to give explicit decision rationale or insights into HTE, which is really crucial for complex disease treatment.
Therefore, we mainly discuss the studies on CATE estimation below.

CATE research generally has two focal points.
One is developing general CATE estimators with favorable statistical properties using statistical methods such as single-index models \cite{song2017semiparametric, FENG2022107554} or (causal) machine learning methods such as causal forest \citep{athey2019generalized} and meta-learners \citep{he2022x, polley2011super, van2011targeted}.
Some studies involve complicated computations or even black-box models \cite{johansson2016learning, yoon2018ganite, alaa2018limits} that are inherently unexplainable to improve estimation accuracy, failing to provide sufficient interpretation catering to clinical practitioners.
The other focus emphasizes identifying and characterizing subgroups with statistically significant and practically meaningful heterogeneity in treatment response, such as \citep{chen2015prim,su2009subgroup}.
Tree-based methods such as \cite{hapfelmeier2018subgroup, lipkovich2011subgroup,seibold2016model,su2009subgroup} have a natural advantage for this end due to their heuristic mechanism on population segmentation.

In particular, there is a growing field of CATE studies with rule ensemble models that follow quite different philosophies and workflows.
As examples, \cite{dwivedi2020stable} proposes a method called Stable Discovery of Interpretable Subgroups via Calibration to discover interpretable and stable rule sets from well-calibrated samples across 18 popular CATE estimators. 
Similarly, \cite{bargaglistoffi2024causal} uses multiple CATE estimators as outcomes to generate tree ensembles for rule generation and then prunes the rules with certain penalties. 
They re-estimate CATE for the subpopulation characterized by the remaining rules. 
\cite{wang2021causal} proposes Causal Rule Sets to identify subgroups with enhanced treatment effect through a generative Bayesian framework that assumes a prior for the actual rule set and a Bayesian logistic regression model to improve it. 
As mentioned earlier, all these studies presume a limited number of true subgroups, which may lead to the truth of complex disease being under-explored, especially for the population outside the identified subgroups. 

\section{Causal Rule Learning}\label{method}
\subsection{Preliminaries}

\subsubsection{Notation and assumptions}
\label{basic setting}

We consider a triplet $(X, A, Y)$ where $X = (x_1, x_2,\dots,x_p)^\top$ denotes a $p$-dimensional vector of sample characteristics (covariates) and 
$A \in \mathcal{A} = \{-1,1\}$ represents the treatment received, with -1 indicating negative treatment (or control) and 1 indicating positive treatment.
Correspondingly, we have $Y(1),Y(-1) \in \mathbb{R}$ denote the potential outcomes for the two treatments, and $Y = Y(A) $ is the observed outcome. 
In this paper, we assume that higher values of $Y$ are preferred. 

Suppose that we have observed a data set $\mathcal{D} = \{X_i, A_i, Y_i\}^N_{i=1}$ which comprises $N$ independent and identically distributed samples. 
This data set may come from an observational study or, ideally, a randomized controlled trial (RCT) setting. 
The proposed framework can be applied to data from both settings but requires the satisfaction of the three basic assumptions of the Neyman--Rubin potential outcome framework \cite{rubin1980randomization,rosenbaum1983central}, namely the Stable Unit Treatment Value Assumption, Unconfoundedness Assumption, and Overlap Assumption. 
Details of these assumptions are given in Appendix \ref{apx:rubin.assumps}.

\subsubsection{ITE based on causal rules (subgroup CATEs)}\label{causal rules}

In this study, we estimate ITE directly through a set of causal rules, by which we mean that the rules are identified from a causal analysis framework.
As a natural way to represent human knowledge and a widely-used tool in clinical guidelines, they uncover variables' potential interaction (synergistic effect) and define certain subgroups to help us understand the heterogeneity of the treatment effect. 
Most importantly, \cite{friedman2008predictive} and \cite{yubinPDR2019} show evidence that rule-based models have comparable and sometimes even better performance than black box methods like neural networks.
Therefore, rules match our needs for both accuracy and interpretability.
This paper formulates a rule as: 
  \begin{equation}
  	r(X) =  \prod_{j=1}^{p} {1}(x_j \in s_j),
  	\label{def:rule}
  \end{equation}
where $x_j$ denotes the $j$-th element of $X$,  $s_{j}$ is a specified subset of all values of $x_j$,  and $\mathop{1}(\cdot)$ is the indicator function yielding 1 when $x_j \in s_j$ and 0 otherwise. 
For example, \textit{$x_1 > 0 $ and $x_2 < 0$ and $x_3 >5$} is a rule representation consisting of three components.  Each component (\eg $x_1 > 0$) has a splitting variable ($x_1$) and a given cut-off value or threshold (0), which defines $s_1$. 
A rule $r(X)$ maps an input $X$ into $\{0,1\}$, representing the satisfaction of a rule with 1 and 0 otherwise. 
Note that a rule is usually represented only by the variables that help to split the whole population.
For those variables that do not contribute to splitting the population, we assume $s_j = \mathbb{R}$. 

As a formulated version of Figure \ref{fig:motivation}, the ITE can be represented by:
 \begin{equation}
  	\tau(X) := \sum_{m=1}^M \beta_m \tau_m r_m(X), 
  	\label{def:true.ite}   
  \end{equation}
where $r_m(X)$, $\tau_m$, and $\beta_m$ denote the satisfaction of the $m$-th rule (0 or 1) given $X$, its corresponding subgroup CATE, and weight in the linear combination, respectively.
Each subgroup CATE makes a different contribution to ITE, as reflected by the sign and magnitude of the weights  ${\beta}$ and the subgroup CATEs. 
More significant absolute values of these quantities imply a greater contribution. 
It should be noted that (\ref{def:true.ite}) only models an additive relationship between the potential subgroups and does not account for more complex non-linear interactions between groups.
We use this assumption mainly for clinical interpretability and simplicity for the demonstration of the CRL framework.
For more complex scenarios, we can easily incorporate potential nonlinear interactions into  (\ref{def:true.ite}) to capture more refined patterns of HTE. 
For instance, $ \tau(X) := \sum_{m=1}^M \beta_m \tau_m r_m(X) + \sum_{i<j} \gamma_{i,j} f(\tau_i r_i(X),\tau_j r_j(X))$ where $\gamma_{i,j}$ is the coefficient for the between-group interaction term between group $i$ and group $j$.

Given above, the problem then boils down to the revelation of $\{r_m(X),\tau_m,\beta_m\}^{M}_{m=1}$.
Various methods can be used to find the rules and estimate the quantities. 
The final choice depends on the particular problem and intended audience. 
The next section shows our CRL framework to achieve these goals.

\subsection{CRL Workflow} 
\label{workflow}

\subsubsection{Rule discovery via causal forest}
  \label{rule discovery}

Our model utilizes causal forests \cite{wager2018estimation} to derive the initial set of potential rules and corresponding CATE estimates. 
Causal forest is a non-parametric method for heterogeneous treatment effect estimation whose estimator has been shown to have good statistical properties like point-wise consistency and asymptotic normality for the true CATE under the three Neyman-Rubin assumptions \cite{wager2018estimation}. 

A causal forest consists of many causal trees \cite{athey2016recursive}, each trained using a random subsample of observations and covariates. 
Given a causal tree, it is easy and intuitive to decompose the tree into multiple rules: the path from the root to a leaf node naturally forms a rule formulated in definition (\ref{def:rule}).
Instances within a leaf node directly make up a potential subpopulation. 
Further, we use the following plug-in estimator of difference-in-means as an estimate for the corresponding CATE of a given subgroup $G$ in a leaf node: 

\begin{equation}
   	\hat{\tau}_G = \frac{\sum_i Y_i {1}(A_i = 1, X_i \in G)}{\sum_i {1}(A_i = 1, X_i \in G)} - \frac{\sum_i Y_i {1}(A_i = -1, X_i \in G)}{\sum_i {1}(A_i = -1, X_i \in G)}.
  	\label{def: subgroup cate estimates}
\end{equation}

\subsubsection{Rule selection with the D-learning method} 
\label{rule selection}

Given the $\{\hat{r}_m,\hat{\tau}_m\}^{M}_{m=1}$ obtained from last step and our linear assumption on the $\tau(X)$ in (\ref{def:true.ite}), we provide our estimation of ITE formulated as
\begin{equation}
    \hat{\tau}(X)=\sum_{m=1}^M\beta_m\hat{\tau}_m\hat{r}_m=\left(\hat{\tau}\circ\hat{r}(X)\right)^\top \beta:=\tilde{X}^\top\beta,
    \label{def:CRL_tau}
\end{equation}
where $\circ$ is the Hadamard product of $\hat{r}(X)=\left(\hat{r}_1(X),\cdots,\hat{r}_M(X)\right)^\top\in\R^{M}$ and
$\hat{\tau}=\left(\hat{\tau}_1,\cdots,\hat{\tau}_M\right)^\top\in\R^M$. 
$\beta = (\beta_1,\beta_2,\cdots,\beta_M)^\top$.
This is merely a transformation from the original sample vector $X$ to the modified rule space $\tilde{X} \in\R^{M} $. 

Due to the randomization and greediness introduced by forest-based methods, fake and redundant rules are inevitably generated \cite{wang2020an}. 
Therefore, the rule selection process aims to filter out these irrelevant and unnecessary rules from the initial rule set through the learning of the coefficients $\beta$ to improve the accuracy of ITE estimation and size down the rule set for interpretability purposes. 
For this end, we leverage the D-learning method \cite{qi2018d} 
to learn a sparse combination of rules that can estimate ITE accurately. 

The original D-learning method aims to directly learn the optimized individual treatment decision $d^*(X)$ in a single step without model specification, defined as:
\begin{equation}
d^*(X) = \textnormal{sign} \Big(\mathbf{E}\Big[\frac{YA}{\pi (A,X)}| X\Big]\Big) = \textnormal{sign} (f_0 (X)),
\label{def:ITR}
\end{equation}
where the $f_0(X):=\mathbf{E}\Big[\frac{YA}{\pi (A,X)}| X\Big]$ is the desired true $\tau(X)$.
(\ref{def:ITR}) leverages the propensity score defined as the probability of receiving treatment $a$ given $X$, \ie $\pi(A,X) = Pr(A|X)$. 
For RCT data, we assume this score is already known.
For observational data, we have to specify a proper model and learn the score from the data.  

It has been demonstrated in  \cite{qi2018d} that under the assumption of interchange between differentiation and expectation, 
\begin{equation}
    f_0(X) \in \argmin_{f} \mathbf{E} \Big[\frac{(2YA-f(X))^2}{\pi(A,X)}| X\Big],
\end{equation}
which transforms the estimation of unobserved true ITE into an expectation maximization problem where $f(X)$ can be any proper form, allowing a flexible specification, whether linear or nonlinear.
Therefore, we replace $f(X)$ with (\ref{def:CRL_tau}) and impose additional $\ell_1$ penalty on $\beta$ to ensure sparsity.
Then the estimation of $f_0(X)$ is reformulated as an $\ell_1$ penalized regression problem, and we can obtain the  least absolute shrinkage and selection operator (LASSO) estimator \cite{tibshirani1996regression} of $\beta$:
\begin{equation}
	\hat{\mathbf{\beta}}=\arg\min_ {\mathbf{\beta}}\Big\{ \frac{1}{2N} \sum^N_{i=1} \frac{1}{\pi(A_i,X_i)} \{2Y_iA_i - \tilde{X}_i^\top \beta\}^2+\lambda \Vert \beta \Vert_1\},
	\label{def:dlearning}
\end{equation}
where $\lambda\geq 0 $ is the tuning parameter.

Just as \cite{yubinPDR2019} states, when we impose sparsity on our model, we limit the number of non-zero parameters in the model and interpret the rules corresponding to those parameters (based on sign and magnitude) as meaningful to the outcome of interest.
Since a large number of weights are zero (\ie $\hat \beta$ usually has many elements equal to 0), their corresponding rules are actually removed from the candidate set and do not contribute to our estimator.  
Imposing this sparsity of rules also highlights the distinction between our CRL estimator and the causal forest estimator, which simply takes the average over all the tree estimates (subgroup CATEs in our method) as the final estimates, while our method learns a weighted combination of those estimates. 
In other words, we reweight the tree estimates instead of using them equally. 
In this sense, causal forest estimates can be viewed as a special case of our method.  We show in Figure \ref{fig:rst.simu}(d) that reweighting the rules with the D-learning method can eliminate a large proportion of unwanted rules that the causal forest generates.

With the learned $\hat{\beta}$,  we can then derive the treatment decision guided by our estimation as $\hat{d}_N(X):=\mathrm{sign}(\tilde{X}^\top \hat{\beta})$.
Further, the expected outcome (or ``value function'' in D-learning) under a given treatment decision $d$ and our decision $\hat{d}_N$ is respectively defined as:
\begin{align*}
     V(d)&=\mathbf{E}[Y|A=d(\mathbf{X})],\\
      V(\hat{d}_N)&=\mathbf{E}[Y|A=\hat{d}_N(\mathbf{X})]
 \end{align*}
where $\mathbf{X}\in\R^{N\times p}$ is the original design matrix composed of $N$ samples and $p$ covariates.
Both $V(d)$ and $V(\hat{d}_N)$ yield an $N$-dimensional vector for every sample.

With the above definitions, we show below that convergence rates of the expected outcome under our treatment decision, \ie $V(\hat{d}_N)$,
can be achieved when the following assumptions and conditions hold:

\begin{enumerate}
    \item \textbf{The compatibility condition.} 
    For some constant $\phi(S)$ and any vector $\beta\in\R^M$ with $\|\beta_{S^c}\|_1\leq3\|\beta_{S}\|_1$, we have
\begin{equation}
\label{eq:compatibility}
\|\beta_{S}\|_1^2\leq\left(\beta^\top\hat{\Sigma}\beta\right)|S|/\phi^2(S),
    \end{equation}
    {where $\beta_S$ denotes the subvector of $\beta$ consisting of the components indexed by the set $S$, $\beta_{S^c}$ is the complement of $\beta_S$,} $\hat{\Sigma}=\frac{1}{N}\tilde{\mathbf{X}}_N^\top\tilde{\mathbf{X}}_N$, and $\tilde{\mathbf{X}}_N=\left(\tilde{X}_1,\cdots,\tilde{X}_N\right)^\top \in\R^{N\times M}$. 
    When the parameter is the oracle $\beta^*$, we use $s_*$ and $\phi_*$ to replace $|S_*|$ and $\phi(S_*)$ where $S_{*}=\{j:\beta_{j}^{*}\neq0\}$ for simplicity.
    
    \item Assume $Y_i=m(X_i)+A_i\delta(X_i)+W_i$, where $m(X_i)$ is the main effect of covariates $X_i$ for both treatments and $W_i$ is the mean zero random error term with its variance $\sigma^2>0$. Let $\epsilon_i=2R_iA_i-\mathbf{E}\left[\frac{R_iA_i}{\pi(A_i,X_i)}\mid X_i\right]$ for $i=1,2,\cdots,N$.
    \item Assume $m(X_i)=X_i^\top\gamma_0$, where $\|\gamma_0\|\leq\mathcal{O}\left(\sqrt{\log(2M)}\right)$.
    \item $\max_{i\in[M]}|\tau_i|\leq a$ for some $a>0$.
    \item There exists some $\rho>0$ such that $\gamma^\top{\Sigma}\gamma\leq\rho\|\gamma\|_2^2$ for any $\gamma$, where $\Sigma=\frac{1}{N}\mathbf{X}_N^\top \mathbf{X}_N$ and $\mathbf{X}_N$ is the original design matrix.
    \item Define the oracle $\mathbf{\beta}^*$ as $\beta^*=\mathrm{argmin}_{\beta}\{\|\tilde{\mathbf{X}}_N\beta-f_0({\mathbf{X}}_N)\|_2^2/N+\frac{4\lambda^2s_{\boldsymbol{\beta}}}{\phi^2(S_{\boldsymbol{\beta}})}\}$, then assume $\|\tilde{\mathbf{X}}_{N}\boldsymbol{\beta}^{*}-f_0({\mathbf{X}}_{N})\|_{2}^{2}/N\leq\lambda\|\hat{\boldsymbol{\beta}}_{S_{*}}-\boldsymbol{\beta}_{S_{*}}^{*}\|_{1}$ where $S_{*}=\{j:\beta_{j}^{*}\neq0\}$, $\hat{\beta}_{j,S_*}=\hat{\beta}_j\mathbb{I}(j\in S_*)$.
    \item \textbf{The general Margin condition (gMC)}. 
For any $\epsilon>0$ there exists some gMC constant $C>0$ and $\alpha>0$, such that $\mathbb{P}(|f_0(X)|\leq \epsilon)\leq C\epsilon^\alpha$.

\end{enumerate}

Specifically, Assumption (1) is a standard condition in the LASSO literature, ensuring that the $\ell_1$-norm of the coefficients on the true support can be controlled by the prediction norm. This property is crucial for establishing oracle inequalities and consistency results (see \cite{vandeGeer2009oracle}). Assumptions (2)--(5) pertain to the modeling of the reward function $Y_i$ and impose restrictions on the original sample space, following \cite{qi2018d}. Assumption (6) is a technical condition introducing an oracle linear parameter $\beta^*$ and describing its approximation behavior to the true decision function $f_0$, as also considered in \cite{qi2018d}. Finally, Assumption (7) characterizes the distributional behavior of the decision function near zero; larger values of $\alpha$ correspond to a larger exponent $1+\alpha/(2+\alpha)$, which in turn yields a sharper upper bound in (\ref{eq:bound}).

Then, we provide the following main theorem:

\begin{theorem}\label{thm}
Let the tuning parameter $\lambda$ be
    \begin{equation}\lambda=16\sqrt{2}t^2\sqrt{\frac{\log^2(2M)}{N}}.\end{equation}
    If the true treatment decision function is linear, that is, $f_0=\tilde{\mathbf{X}}_N{\beta}^*$, then when the assumptions (1)-(7) hold, 
    with probability at least $1-\frac{C}{t^2}$ where $C$ depends on $a,\rho, \sigma$, we have 
    \begin{equation}\label{eq:bound}
        \|V(d^*)-V(\hat{d}_N)\|_2\leq C_2\left(\frac{\log(2M)}{N}\right)^{\frac{1+\alpha}{2+\alpha}},
    \end{equation}
    where $C_2$ is determined by the gMC constant, $t$, $\phi_{S_*}$, and $|S_*|$.
\end{theorem}

\begin{remark}
This theorem provides the explicit convergence rate of the value difference to zero when the true ITE function is indeed linear in the transformed sample space, i.e., $f_0 = \tilde{\mathbf{X}}_N \beta^*$. The rate $(\log(2M)/N)^{\frac{1+\alpha}{2+\alpha}}$ depends on the number of candidate rules $M$ only through its logarithm, justifying the use of a large initial rule pool generated by the causal forest. The exponent $\frac{1+\alpha}{2+\alpha}$, governed by the gMC constant $\alpha$, highlights the beneficial effect of a margin condition—stronger separation between treatment effects (larger $\alpha$) leads to faster convergence of the policy value. This theoretical guarantee reinforces the practical utility of the CRL framework.
The proof of Theorem \ref{thm} can be found in Appendix \ref{apx: theory}.
\end{remark} 

\subsubsection{Rule analysis}
\label{rule analysis}
Interpreting our estimated \taux relies on understanding the subgroups contributing to it. 
A comprehensive analysis of the subgroups identified is what needs to be improved in existing studies. 
This analysis is indispensable since i) we cannot guarantee that all the rules D-learning selects have their stated impact on our estimators, and ii) the size of the rules may still be too large to achieve human comprehension. 
Thus, we show a procedure to further analyze the rules remaining in the candidate set to find a relatively smaller yet more promising subset of rules. 
For clarification, this procedure continues to remove a large number of rules, primarily to enhance interpretability, and the estimation of ITE is still made on all the rules D-learning selects (rules with non-zero weights). 
This procedure includes but is not necessarily limited to the following three parts: overall analysis, rule significance analysis, and rule decomposition analysis. 
Unlike previous studies, our analysis does not focus on the subgroups with higher treatment effects but on those with significantly different ITE estimates, whether higher or lower.

\textbf{Overall analysis}. 
Suppose D-learning selects a rule set that consists of $n_1$ rules ($n_1\ll M$) whose weights are non-zero in model (\ref{def:dlearning}), \ie $\hat{R}_1 =\{\hat{r}_1, \hat{r}_2,\dots,\hat{r}_{n_1}\}$ (for simplicity, we use $\hat{r}_j$ instead of $\hat{r}_j(x)$ to represent a rule). 
The corresponding CRL estimates of ITE $\hat{\tau}_{CRL}$ are also obtained based on this rule set.
As estimation accuracy is the basis of interpretability, the overall analysis aims to evaluate how this rule set collectively performs on ITE estimation using various performance metrics.
For simulated data where the true treatment effect is already known, we define Mean Squared Error (MSE), Mean Potential Outcomes (MPO), and Population Overlap (PO) to evaluate and compare our estimator with other baselines.
For observational data where the ground truth can never be observed, the above metrics can no longer be used, so we use ITE-based prediction accuracy, Empirical Expected Outcome (EEO), Mean-squared Prediction Error (MSPE) of transformed outcome, and Treatment Efficient Frontier.
Details of these metrics can be found in section \ref{simulation} for simulated data and \ref{realworld.metrics} for real-world data.

Suppose the above overall analysis shows that our method is able to give comparable ITE estimates to those of baseline methods, we then continue the following analysis, which evaluates each rule in terms of i) the ability to distinguish different levels of ITE, \ie rule significance, and ii) the role of each component in the rule, \ie rule decomposition analysis.

\textbf{Rule significance analysis}. 
A rule defines two subpopulations: the population that satisfies the rule $G^Y=\{i|r(X_i)=1\}$ and the rest that does not 
$G^N=\{i|r(X_i)=0\}$. 
Therefore, we want to test the discriminating power of a rule by checking if $G^Y$ and $G^N$ differ significantly in the magnitude of ITE estimates, using the well-known two-sample Kolmogorov--Smirnov test \cite{berger2014kolmogorov}. 
This test is commonly used to determine whether two samples originate from the same continuous distribution. 
In our context, the null hypothesis posits that the ITE estimates for groups $G^Y$ and $G^N$ derive from an identical distribution. If the null hypothesis cannot be rejected, we assume there is no significant difference in the ITE estimates between the two groups.
Specifically, for a given significance level $\alpha$, a $p$-value $\leq \alpha $ implies that the rule can identify a subgroup with significantly different treatment effects than the rest of the population.
We perform this test on each rule in $\hat{R}_1$ and remove those with $p$-value $>\alpha$. 
To avoid a high false discovery rate (FDR) in this multiple testing, we recommend using FDR control techniques such as the well-known Benjamini-Hochberg method \cite{benjamini_controlling_2018} and removing rules based on these FDR-adjusted $p$-values instead of the original ones. 
The choice of $\alpha$ and the FDR level $q$ is up to domain experts and should be justified rigorously. 
In our experiments below, we simply adopt the conventional choice of 0.05 for both thresholds for illustrative purposes.
The remaining $n_2$ rules ($n_2\leq n_1$) form the new candidate rule set $\hat{R}_2$.

\textbf{Rule decomposition analysis}. 
Decomposition analysis evaluates how each component in a rule contributes to the complete rule in differentiating the treatment effect. 
For example, if we have a rule $\hat{r_j}$ that has three components $a$, $b$ and $c$, we can derive three revised rules: $\hat{r}^a_j$, $\hat{r}^b_j$ and $\hat{r}^c_j$. 
$\hat{r}^a_j$ is the rule with component $a$ removed from the original rule, that is, the rule formed only by components $b$ and $c$, and similarly for $\hat{r}^b_j$ and $\hat{r}^c_j$. 
We do the above significance analysis on all revised rules and obtain their corresponding $p$-values $p^a_j$, $p^b_j$, and $p^c_j$. 
We think a higher $p$-value of the revised rule than that of the original complete rule, say, $p^c_j>p_j$, implies that the revised rule $\hat{r}^c_j$ is less significant in distinguishing subpopulations with different levels of ITE estimates than the original entire rule $\hat{r_j}$. 
Hence, the removed component $c$ contributes to the significance of the ITE difference.  
The higher the $p$-value of the revised rule, the more critical the variable being removed is, which reveals the variable most accountable for the group's heterogeneity. 
In contrast, a lower value suggests the corresponding component(s) may be unnecessary and meaningless to the complete rule since their removal does not reduce the significance of the difference but increases it. 
Therefore, if all revised rules have higher $p$-values ($p_j <$ min$\{
p^a_j, p^b_j, p^c_j\}$), then all the components in the rule work synergistically to contribute to its discriminating power. 
In this step, we retain the rules that have higher $p$-values on all the corresponding revised rules to form a new refined set $\hat{R}_3$ with $n_3$ rules ($n_3 \le n_2$). 

In conclusion, the rule analysis step shows a universal way to analyze the identified rules through different levels. 
However, these tools highly rely on the data at hand. 
Therefore, we highly recommend using this analysis procedure as a screening step for the lengthy list of rules and picking out the top subgroups for more rigorous statistical tests and further interpretation.    

\subsection{Discussion on the workflow design of CRL}

As presented in the last section, our whole workflow helps us to estimate and build a better understanding of HTE for complex diseases by finding potential subgroups of heterogeneity and exploring the relationship between ITE and  CATEs of these subgroups.
As Figure \ref{fig:workflow} shows, the rule discovery step first identifies a large pool of interpretable rules as candidate subgroups.
The rule selection process then removes a great proportion of non-informative and fake subgroups from the candidate set and uses the remaining $M$ subgroups as the selected sources of effect heterogeneity to learn the connection between ITE and subgroup CATEs (through $\tau$ and $\beta$).
Finally, the rule analysis part evaluates how accurate these $M$ subgroups estimate ITE as a whole and looks deeper into each of the $M$ subgroups to evaluate how each subgroup and the factors involved in the subgroup contribute to HTE estimation.

This modular, three-stage architecture of CRL is a deliberate design choice that prioritizes transparency, controllability, and clinical practicality, which are often more critical and fundamental than efficiency for generating trustworthy and actionable insights that can be integrated into clinical decision-making processes \cite{kundu2021ai,shaban2020explainable}.
Our workflow ensures that the entire process is interpretable and auditable, as each stage produces inspectable intermediate results. 
More importantly, it provides domain experts with multiple opportunities to intervene and guide the analysis, transforming the workflow from a black box into a collaborative, human-in-the-loop system. 
For instance, experts can pre-configure parameters (e.g., adjusting \textit{min.node.size} to influence rule complexity) in the discovery phase, manually curate the initial candidate rule pool based on clinical knowledge after discovery, extend the ITE assumption to include non-linear interaction terms if theorized, and flexibly choose the depth of analysis applied to the selected rules.

Primarily due to the uncertainty and randomization of rule generation via causal forest, there may be multiple combinations of $\{\hat{\tau}_m, \hat{r}_m\}^M_{m=1}$ generated in the rule discovery step, which may yield equivalent estimations of ITE.
In this sense, our model is not identifiable.
However, this problem is mitigated to a great extent by the subsequent selection step and analysis step.
Besides, we can fit CRL repeatedly across resampled or perturbed datasets and identify rule combinations that consistently appear.
This operation is recommended by \cite{Bin2020Veridical,rewolinski2025pcs}, which emphasizes the importance of reproducible and stable results under data and model perturbations.
The results of our simulation studies in section \ref{evaluation} also verify its effectiveness. 
If, after such stability assessments, multiple rule combinations persist, this may reflect a genuine multiplicity in the underlying data-generating process and the inherent complexity of treatment effect heterogeneity. 
In such cases, the non-uniqueness of rule combinations is not a flaw of our method, but rather a strength. 
Our model thus serves as an exploratory tool for the complexity of HTE, and its final interpretation and validation should be guided by domain experts with prior knowledge and rigorous randomized trials.

\section{Evaluation of CRL Performance}
\label{evaluation}

In this section, we use both simulated data and real-world data to evaluate how well CRL performs in estimating and providing insights to HTE. 
Specifically, the simulated RCT data is mainly used to evaluate the estimation accuracy of CRL, while the real-world application demonstrates how to apply CRL in observational data and how CRL gives insights to us on real-world problems (enhanced understanding).
In addition, we also compare CRL with several previously mentioned baseline methods, namely outcome weighted estimation (OWE) \cite{zhao2012estimating}, causal tree (CT) \cite{athey2016recursive}, causal forest (CF) \cite{wager2018estimation}, patient rule induction method (PRIM) \cite{chen2015prim} and causal rule ensemble (CRE) \cite{bargaglistoffi2024causal} in both the simulation experiments and real-world applications. 
Baseline methods are implemented with R package \textit{personalized} \cite{R_personalized} for OWE (linear), \textit{causalTree} \cite{causalTree_pkg} for causal tree, \textit{subgrpID} \cite{subgrp2017huang} for PRIM, \textit{grf} \cite{tibshirani2023package} for causal forest.
Code is available at \url{https://github.com/yingwu2017/Causal-Rule-Learning/.}

\subsection{Simulation study one: the overall performance of CRL}
\label{simulation}

\subsubsection{Data setups}

To evaluate how well CRL performs in estimating HTE, we first generate simulated RCT data sets $\{X_i, A_i, Y_i\}_{i=1}^N$ based on assumption (\ref{def:true.ite}) with varying configurations. 
Specifically, we generate covariates $X \in \mathbb{R}^9$ where the first three elements (features) $x_1,x_2$, and $x_3$ are randomly drawn from $\{0,1\}$ with equal probability ($x_3$ is sampled differently for some setups of the true group, shown later). 
The last six features are
$\{x_j| j = 4,5,\dots,9\} \overset{\mathrm{i.i.d}}{\sim} \mathcal{N}(0,2)$.
For simplicity, we ignore the sample subscript notation $i$, and the remaining subscript represents certain features unless otherwise noted. 
The baseline effect is defined as
$f(X ) = 1+x_1^2+3x_1x_6 + 0.4x_4x_7+x_5-
x_6+0.5x_3x_8$. 
{Based on} (\ref{def:true.ite}), we can define any number of true subgroups and their related true CATEs via a rule form (shown later). 
Once we set the true $\tau(X)$, we define the potential outcomes for both the negative and positive treatments: $Y(-1) = f(X) + \epsilon$ and $Y(1) = Y(-1) + \tau(X)$, where $\epsilon \sim \mathcal{N}(0,1)$. 
We simulate the simplest RCT assignment of the treatment arm: $A_i$ is sampled from $\{-1,1\}$ with equal probability. 
Therefore, the observed outcome in our data set is $Y_i = 2^{-1}(1 + A_i) Y_i(1) + 2^{-1}(1-A_i)Y_i(-1) $.

\begin{figure}
	\centering
	\includegraphics[scale=0.6]{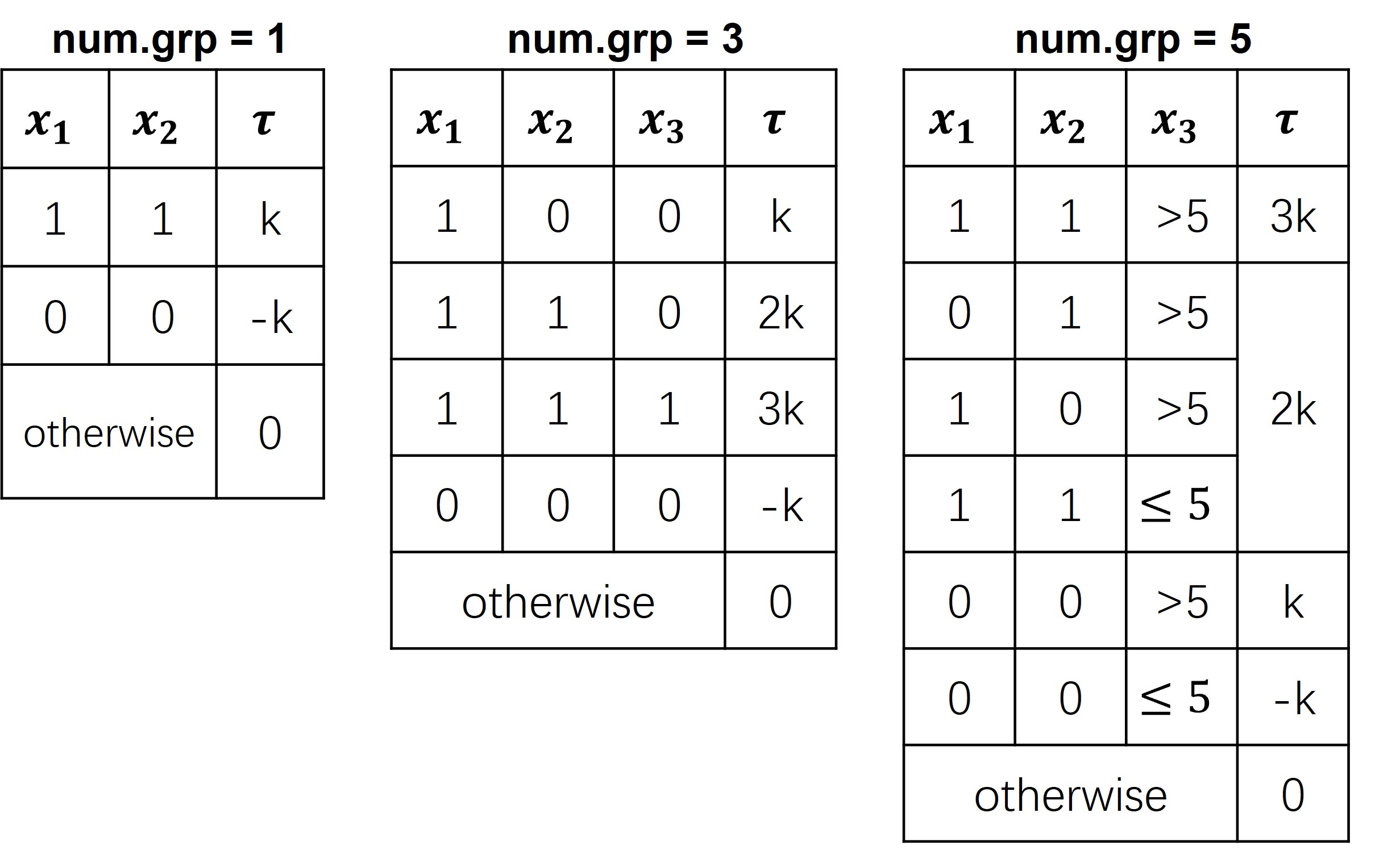}
	\caption{Specification of true treatment effect $\tau(X)$ for varying numbers of subgroups. 
$k$ is the effect size base quantity and $\tau$ is the true treatment effect. 
For each scenario, once $k$ is given, we set $\tau$ to different multiples of $k$ depending on the covariates.
 E.g., for num.grp = 3,   $\tau = k$ if $x_1=1, x_2 =0$ and $x_3 = 0$ whereas $\tau = 2k$ if $x_1=1, x_2 =1$ and $x_3 = 0$.   
 }
\label{fig:num.grp}	
\end{figure}

To sum up, the observed data set $\{X_i, A_i, Y_i \}^N_{i=1}$, together with the predefined baseline effect $\{f(X_i)\}^N_{i=1}$ and true ITE $\{\tau(X_i)\}^N_{i=1}$  forms the basis of the ground truth of the data generation. 
To avoid randomness and to test how the performance of CRL changes with the data settings, we introduce a few variations to our simulated data:  

\begin{itemize}
		\item Number of the observations (num.obs) $N \in \{2500,5000,10000\}$,
		\item  Effect size base quantity $k \in \{5,10,15, 20,25,30\}$, where $k$ is used in the definition of true subgroups and related $\tau$ in Figure \ref{fig:num.grp}.
		\item Number of true positive subgroups (num.grp) defined in rules: $M^* \in \{ 1, 3, 5 \}$. If $M^* = 5$,  $x_3 \sim U(0,10)$, otherwise, it still follows the before-mentioned Bernoulli distribution. 
For a clearer performance comparison, we focus on identifying subgroups whose true treatment effects are positive (\ie true positive subgroups), which most baseline methods only care about. However, we keep the pre-defined negative groups in our setups. 

\end{itemize}
	
Together, we have $3\times3\times6 = 54$ data sets. 
We first make 100 resampled repetitions for each data set and randomly split each repetition into a 70\% training set and a 30\% test set. 
Then, we apply the overall CRL workflow as well as other baseline methods on the training data and report performance on the test data, averaging over 100 repetitions. 

\subsubsection{Performance metrics for method evaluation}

Performance metrics used in this study focus on how accurately a method estimates ITE. 
To evaluate so, we use Mean Squared Error (MSE), \ie $MSE(\hat{\tau}) = N^{-1}\sum^N_{i=1} (\hat{\tau}(X_i)-\tau(X_i))^2$, where $\hat{\tau}(X_i)$  and $\tau(X_i)$ are a certain estimate and the true ITE of instance $i$ , respectively. 
The smaller the value, the better an estimator does. 
Besides MSE, we also evaluate performance from two other supplementary perspectives, namely Mean Potential Outcomes (MPO) and Population Overlap (PO). 

MPO, defined as $N^{-1} \sum^N_{i=1} {Y_i(d(X_i))}$, measures how good the average outcomes can be under a certain treatment assignment strategy $d(X)$.
The larger the value of MPO, the better the treatment strategy is. 
For an intuitive comparison, we use the simplest strategy that the treatment recommendation is purely based on the sign of  ITE estimates, \ie $d(X_i)=1$ if $\hat{\tau}(X_i)>0$ and $-1$ otherwise. 

We define $PO ={|T\cap I|}/{|T \cup I|} \in [0,1]$ where 
$T = \{i \in N| \tau(X_i) > 0\}$ and $I =  \{i \in N|\hat{\tau}(X_i) >0\}$ represent respectively the sets of instances whose true ITE and estimated ITE are positive and $|\cdot|$ is the number of instances in a set. 
This measure evaluates the degree of intersection of the two sets.
When $T \cap I = \emptyset$, $PO = 0$ and when $T = I$, $PO=1$.
Larger values are preferred.

We also use the above metrics for tuning the parameters involved in CRL and other methods.
Please see Appendix \ref{apx:para_tune} for details.

\subsubsection{Simulation results}
\begin{figure}
	\centering
	\includegraphics[width=1\linewidth]{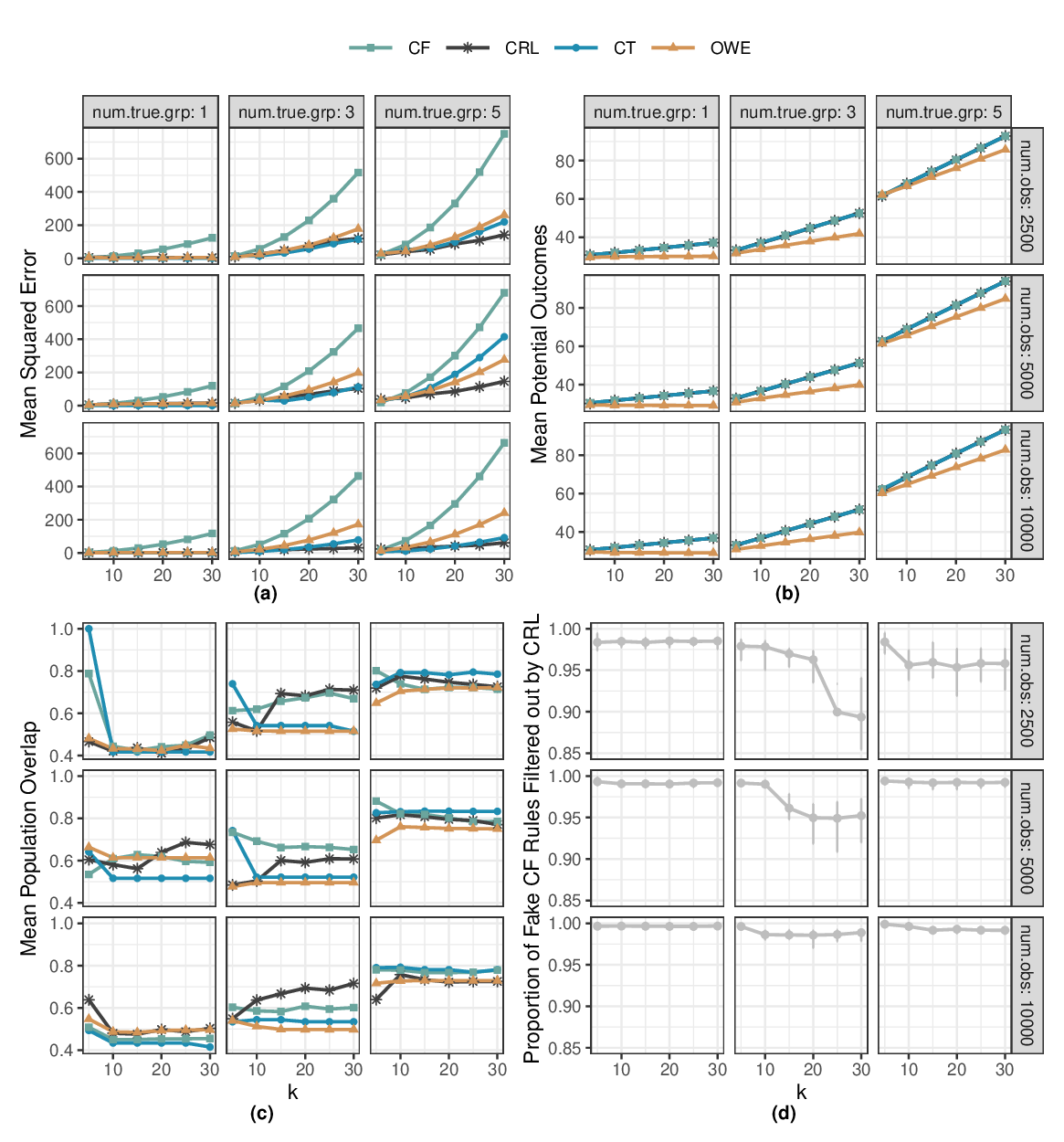} 
	\caption{Performance comparison of CRL (black asterisk) and baseline methods CT (blue circle), CF (green square), OWE (yellow triangle) applied on the simulated data. 
    All performance metrics are averaged over 100 repetitions for each data set. 
    (a) Mean-squared error of treatment effect. 
 (b) Mean potential outcomes.  Note that CRL, CT, and CF have identical performance on MPO, hence their curves overlap. (c) Mean population overlap. (d) Proportion of fake CF rules filtered out by CRL.}
	\label{fig:rst.simu}	
\end{figure}

\begin{figure}
	\centering
	\includegraphics[width=1\linewidth]{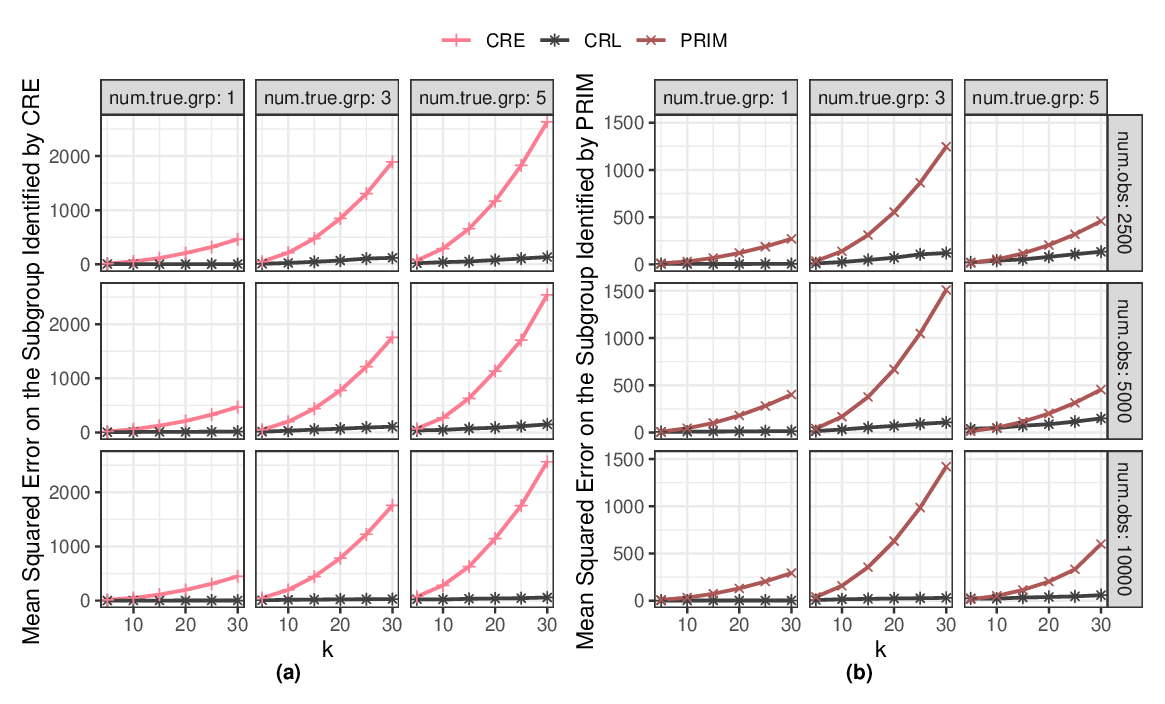}
	\caption{Comparison of mean squared error of treatment effect on specific subpopulations between CRL and baseline methods CRE (pink cross) and PRIM (red X) applied on the simulated data. 
    Each value shown is the average over 100 repetitions for each data set. 
    (a) MSE of CRL versus CRE on the subgroup identified by CRE.
 (b) MSE of CRL versus PRIM on the subgroup identified by PRIM. }
	\label{fig:rst.simu2}	
\end{figure}
This section shows the results of the performance comparison between CRL and other baseline methods for each of the 54 data sets (settings).  
Since the baseline methods vary in research focus and model outputs, not all methods can be directly compared.
Specifically, CT, CF, OWE, and CRL are able to give ITE estimates for every instance in the data, so we compare and show their estimation accuracy on the whole population in Figure \ref{fig:rst.simu}.
For CRE and PRIM, which only identify specific subpopulations with enhanced treatment effect, we compare them with CRL only on the subgroups they respectively identify, shown in Figure \ref{fig:rst.simu2}

Figure \ref{fig:rst.simu}(a) shows the comparison of MSE between CT (blue), CF (green), OWE (yellow), and CRL (black). 
For panels from the top to bottom, the number of observations of the data set varies from 2,500 to 10,000, and from left to right, the number of true positive subgroups varies from one to five. 
Effect size ($k$) is reflected on the x-axis, and each value shown is an average of 100 runs on the specific setting. 
Figure \ref{fig:rst.simu}(b), \ref{fig:rst.simu}(c), \ref{fig:rst.simu}(d), \ref{fig:rst.simu2}(a) and  \ref{fig:rst.simu2}(b) are presented in the same way as Figure \ref{fig:rst.simu}(a) for other metrics.

Figure \ref{fig:rst.simu}(a) shows that each method's MSE increases as the effect size grows across different configurations.  
In general, CRL is almost always among the best performers across all settings. 
For cases where $num.true.grp=1$, CT, CRL, and OWE have almost the same performance. 
Hence, their curves overlap.
Obviously, the superiority of CRL increases gradually as the number of observations and true subgroups increase under large effect sizes  (\eg $k \in \{20,25,30\}$).
Besides, from  Figure \ref{fig:rst.simu2}, we see that no matter for the subgroups identified by CRE (pink) or PRIM (red), CRL has a consistently much lower MSE, indicating that CRL performs better for treatment effect estimation of complex diseases that have many potential contributing factors and interactions between these factors.
In contrast, both CRE and PRIM identify a limited number of subgroups (most of the time, they only identify a single subgroup) and variables, failing to achieve the goal.

Figure \ref{fig:rst.simu}(b) compares how well the involved methods do in treatment recommendations using MPO.
CRL, CT, and CF show almost the same good performance on MPO and outperform OWE in all settings. 
The identical performance of the former three methods is a result of our treatment assignment strategy: We recommend treatment solely on the sign of ITE estimates. 
Regardless of how these methods differ in the magnitude of the estimates, they yield identical treatment recommendations as long as their estimates have identical signs, hence the same values on MPO. 
However, in practice where decisions incorporate cost-benefit analysis or patient prioritization based on the expected magnitude of benefit (as reflected by the ITE magnitude), the superior accuracy of CRL's ITE estimates (Figure \ref{fig:rst.simu}(a)) would be critical for developing more nuanced and effective strategies.

Figure \ref{fig:rst.simu}(c) shows that CRL has comparable performance to baseline methods concerning PO across most settings  (with higher overlap value and lower variance), indicating that it can recover the majority of populations with positive treatment effects.

Since CRL filters out rules from the original rule set generated by CF, we also compute the proportion of CF rules filtered out by CRL, \ie $1-{W_{CRL}}/{W_{CF}}$ where $W_{CRL}=|\{r_m| r_m \in I_{CRL}, r_m \notin T\}|$ represents the number of fake rules in the rule set ($I_{CRL}$) identified by CRL, and $W_{CF}$ represents the same for CF. 
The bigger the value, the better CRL filters out fake rules. 
As indicated in Figure \ref{fig:rst.simu}(d), through the additional D-learning selection and rule analysis procedure, CRL can filter out a high proportion of fake rules CF generates.

At the end of this section, we remind the reader that all variables in the simulated data used herein are independently sampled. To evaluate the performance of CRL under more realistic conditions where variables exhibit correlations, we generate multiple datasets with varying degrees of correlation to test the robustness of CRL's performance in the overall analysis and the p-values and their interpretation in the rule decomposition analysis under correlated settings. 
Results from both experiments confirm that CRL exhibits strong robustness in both performance metrics and rule interpretation, even under significant variable correlations.
For detailed results and discussions, please refer to Appendix \ref{app:correlated_data simulation}.

\subsection{Simulation study two: Effectiveness of D-learning in rule selection}

In CRL's workflow, the D-learning method plays a core role by selecting a sparse and refined rule set from a candidate rule set that includes the true rules and fake rules with corresponding subgroup CATEs, which greatly enhances the generalization and interpretability of the proposed method. 
Therefore, this study mainly tests D-learning from two aspects:
i) The adaptability of D-learning in the rule selection step. 
Precisely, is D-learning able to adaptively select an appropriate number of rules according to the underlying treatment heterogeneity and data complexity? 
We expect the trend that when the number of true subgroups ($M^*$) increases with the number of selected (non-zero) rules.
ii) The accuracy of D-learning in the discovery of true rules, \ie how many true rules can be recovered via D-learning? 
Higher values are preferred.

\subsubsection{Data setups}
Instead of generating explicit covariates and designing specific ground-truth rules as we do in study one, this study starts with directly generating the data fed to the D-learning method, \ie  $\{r_m,\tau_m\}^{M}_{m=1}$ and $\{A_i, Y_i\}^{N}_{i=1}$ where $M$ is the number of candidate rules and $N$ the number of observations.
Specifically, for each simulated individual, we sample $\{r_m\}^{M}_{m=1}$ from $\{0,1\}$ with probability $\mathbf{P}_m \sim \operatorname{Unif}(0.1,\ 0.9)$.
Then $\tau_m$, the subgroup CATE corresponding to a rule, is set to a fixed effect size quantity $k$ for all the true rules, and sampled from $\mathcal{N}(0,1)$ for all fake rules.
To fully demonstrate the performance of D-learning under different data configurations, $k$ is set to 5,10,15, 20,25, and 30, $M$ to 100,300, and 500, and $N$ to 2500, 5000, and 10000.
Accordingly, we specify the first 10, 30, and 50 rules to be the true rules; hence, $M^*$ varies from 10, 30, and 50. 
For simplicity, we pre-define the coefficient in the true linear combination, \ie  $\beta_m = 1/M^*$ for all the true rules and $\beta_m = 0$ for the fake rules.
Then we can define the true ITEs based on \ref{def:true.ite} and sample $\{A_i, Y_i\}^{N}_{i=1}$ in the same way in study one.
 
\begin{figure}
	\centering
	\includegraphics[scale=.85]{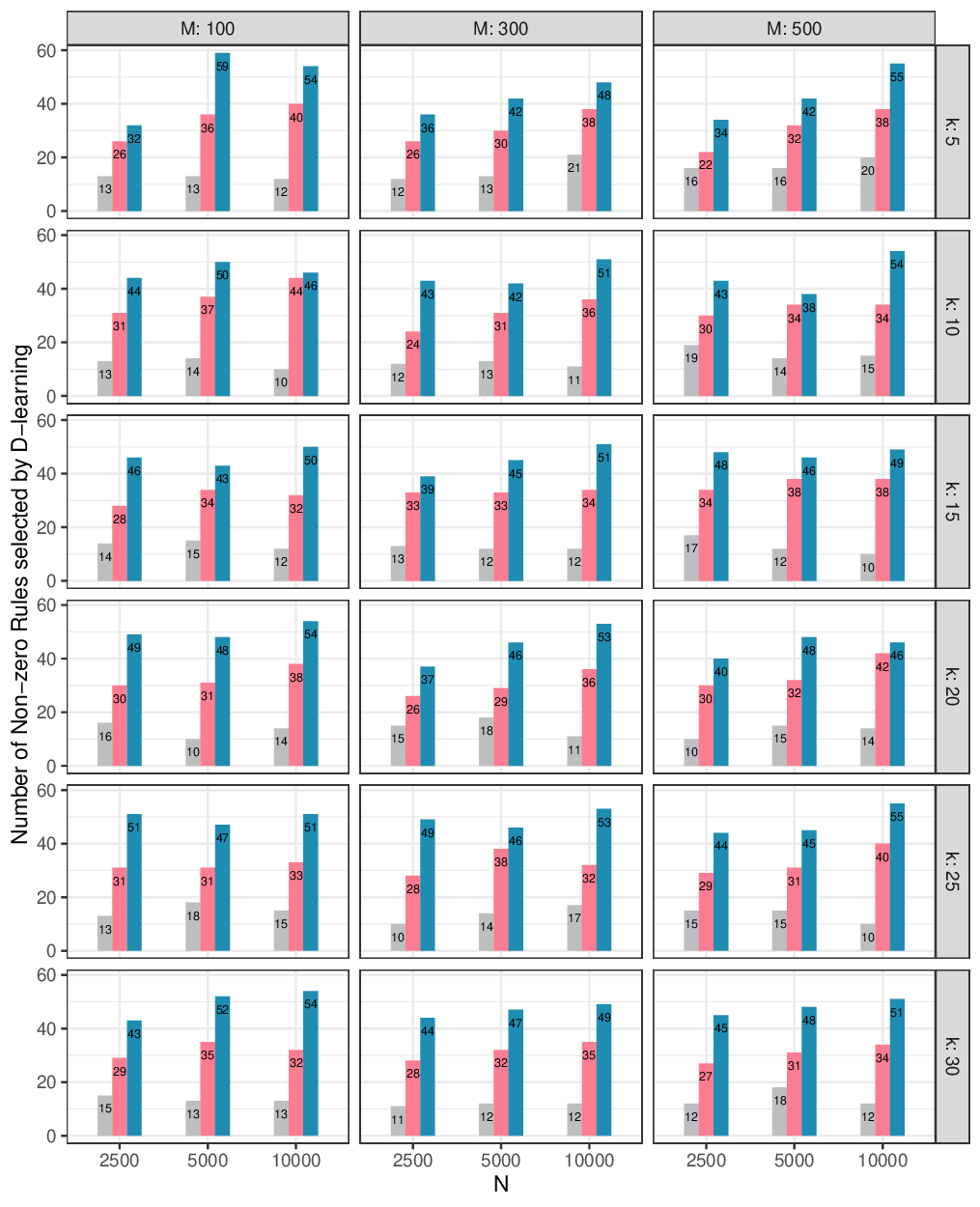}
	\caption{Number of non-zero rules selected by D-learning.
    Each subplot shows the result under different combinations of $\{k,M\}$ and is grouped by the number of observations $N$.
    The gray, pink, and blue pillars show the numbers for $M^*=$ 10, 30, and 50, respectively.}
	\label{fig:study2.1}	
\end{figure}

\begin{figure}
	\centering
	\includegraphics[scale=.85]{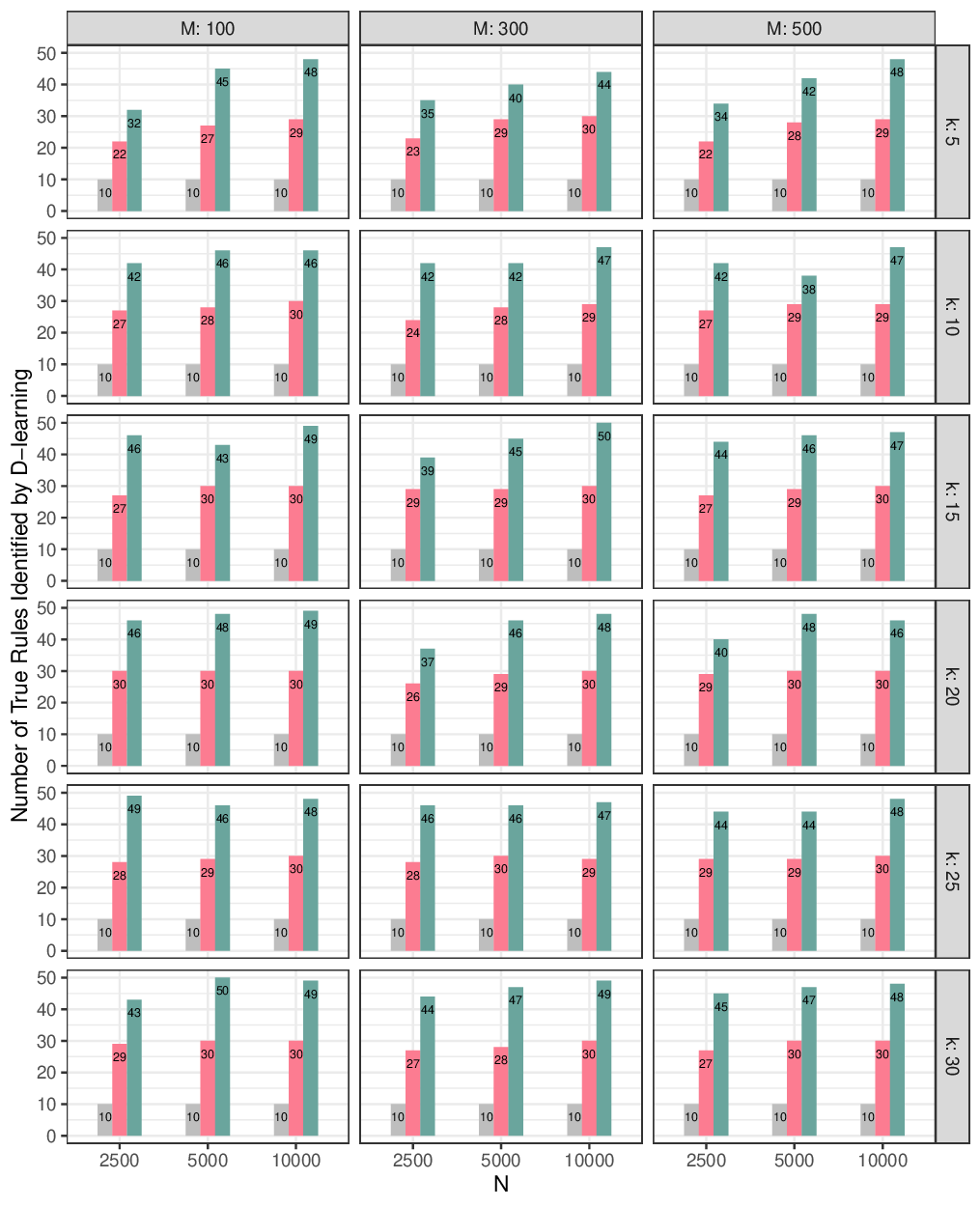}
	\caption{Number of true rules identified by D-learning.
    Each subplot shows the numbers under different combinations of $\{k,M\}$ and is grouped by the number of observations $N$.
    The gray, pink, and green pillars show the results for $M^*=$ 10, 30, and 50, respectively.}
	\label{fig:study2.2}	
\end{figure}

Together, we have $3\ (N) \times 6\ (k) \times 3\ (M^*) \times 3\ (M) = 162$ datasets in total.
We apply the D-learning method on each dataset, and then compare the number of non-zero rules selected under different $M^*$ for 54 different combinations of $\{N,k,M\}$ to see if the aforementioned trend exists.

\subsubsection{Simulation results}

Figure \ref{fig:study2.1} and \ref{fig:study2.2} show the number of non-zero rules and true rules identified by D-learning, respectively.
Clearly, across all settings, the blue pillars in \ref{fig:study2.1} are higher than the pink ones and the pink than the gray, suggesting that the number of non-zero rules selected has an increasing trend with the number of true rules.
Additionally, for most datasets, the number of non-zero rules identified by D-learning consistently approximates and slightly exceeds the number of true rules. 
As evidenced in \ref{fig:study2.2}, the method reliably captures most true rules while retaining a minimal set of redundant ones.

\subsection{Computational complexity and efficiency of CRL}
\label{computational analysis}

Our CRL is a multi-stage framework based on rules, and its computational cost mainly lies in the discovery, selection, and analysis of rules. 
Since the rule analysis step is optional and highly affected by the choice of human experts, our analysis focuses on the first two steps. 
Besides, we do not consider multiple runs of CRL and data splitting here in our analysis, \ie we consider the case that all observations are used for training in both steps.
In the rule generation step, suppose we use $N$ observations with $p$ covariates to train a causal forest with  $T$ trees,  \textit{mtry} $=\sqrt{p}$ and the \textit{min.node.size} $=s$ respectively, then we have a cost of $\mathcal{O}(T \cdot N \cdot \log(N/s) \cdot \sqrt{p})$.  
Subsequently, extracting rules from all the $M$ leaf nodes ($M \approx N/s$) yields a cost of $\mathcal{O}(T \cdot M \cdot \log M )$ and estimating their subgroup CATEs $\mathcal{O}(T \cdot N )$.
In the rule selection step, D-learning selects a sparse subset of rules from the candidate set of size $M$, which typically yields a cost of $\mathcal{O}(M \cdot N \cdot n_1 )$ where $n_1 \ll M$ is the number of non-zero rules selected.
Together, we have a total computational cost of $\mathcal{O}(T \cdot N \cdot \log(N/s) \cdot\sqrt{p} + T\cdot N^2 \cdot n_1/s )$.
The $N^2$ term may enhance the complexity of computation, but in practice, we can mitigate this with a larger $s$, which at the same time yields shorter and thus more interpretable rules.

To better show the computational efficiency of CRL in real practice,
we test CRL's actual CPU runtimes with the data we generated in study one in section \ref{simulation}.
In detail, we test three data settings: $\{N=2500,num.grp=1,k=10\}$, $\{N=5000,num.grp=3,k=20\}$ and $ \{N=10000,num.grp=5,k=30\}$. 
In addition to the original setting of 9 covariates, we generate extra covariates with $p = 18, 81$ where the newly added covariates are all drawn from $\mathcal{N}(0,2)$. 
Therefore, we have 9 datasets in total with $N$ varying from 2500 to 10,000 and $ p$ from 9 to 81.
For each data set, we run CRL 100 times and report the running time quantiles.  
The parameters involved in CRL are fixed to their best values given in Appendix \ref{apx:para_tune}. 
All computations are performed on an Apple M2, 10-core CPU, 16 GB RAM platform using R version 4.4.1.
\begin{table}[h]
    \centering
    \caption{Running time quantiles of CRL (in seconds) based on 100 runs. 
    $N$ denotes the number of observations and $p$ the number of covariates.}
    \begin{tabular}{lccccccc}
        \hline
        $N$    & $p$ & Min & Q1  & Mean & Median & Q3  & Max \\ 
        \hline
   
               2500& 9&    0.34&    0.40&       0.42&        0.42&    0.45&    0.46\\
& 18&    0.37&    0.43&       0.45&        0.45&    0.48&    0.49\\
               & 81&    0.96&    0.99&       1.03&        1.03&    1.06&    1.15\\
               5000& 9&    1.74&    1.77&       1.80&        1.80&    1.83&    1.91\\
         & 18&    2.25&    2.31&       2.40&        2.37&    2.46&    2.66\\
               & 81&    3.19&    3.79&       3.90&        3.95&    4.15&    4.35\\
        10000  & 9         &    4.35&    4.41&       4.56&        4.47&    4.73&    4.93\\
               & 18        &    4.99&    5.37&       5.54&        5.49&    5.73&    6.53\\
               & 81&    5.29&    5.42&       5.75&        5.57&    5.80&    6.78\\ 
               \hline
    \end{tabular}
    
    \label{tab:running time}
\end{table}

As shown in Table \ref{tab:running time}, generally, the results are consistent with our theoretical analysis.
Specifically, running time increases more than linearly with sample size $N$, as when $N$ grows from 2500 to 10,000, the increase in time is over four times ($4.35 \approx 0.34 \times 12$).
It also increases with the number of covariates $p$. 
For example, when $p$ grows from 9 to 81, the increase in time is nearly about three times ($0.96 \approx 0.34 \times 3$).
Despite the $N^2$ term in the theoretical cost, the actual running times remain practical (under 7 seconds even for the largest dataset), confirming that CRL is computationally feasible for the data scales commonly encountered in clinical research.

\subsection{Real-world data application}
\label{realword}
In this part, we apply the CRL workflow to atrial septal defect (ASD) data to demonstrate its usage and performance (especially interpretability) in real-world observational data. 
ASD is a congenital heart defect characterized by a hole in the wall (atrial septum) that separates the upper chambers (atria) of the heart. 
As a common type of birth defect, ASD has a prevalence of 2.5 out of 1000 in live births and 25\% to 30\% present in adulthood \cite{brida2021atrial,vanderlinde2011birth}.

In this study, we focus on two common ASD treatments: percutaneous interventional closure (PIC) and minimally invasive surgical closure (MISC). 
PIC involves using catheters to reach the heart through blood vessels and repair the defect with an occluder, while MISC involves making a small incision in the chest, allowing access to the heart for patching.
We investigate the HTE of PIC versus MISC (negative treatment) on hospital-free days within one year after discharge (HFD1Y).
Hospital-free days are the days patients spend outside the hospital. 
Instead of \textit{survival} or \textit{days alive} that overemphasize the survival goal, this metric is more pragmatic and patient-centered, reflecting more information on the patient's quality of life \cite{auriemma2021hospital}. 

In what follows, we first briefly introduce the ASD data. 
Then, we list a few performance metrics designed particularly for HTE estimation in real-world data where the ground truth of the treatment effect and the treatment assignment mechanism are unknown. 
Finally, we show the performance comparison of CRL and other baseline models on estimation accuracy and enhanced understanding of ITE.
Note that all the CRL rules mentioned and presented in the following text have already been screened through the rule analysis process.

\subsubsection{Overview of the ASD data}

The ASD data used in this study were extracted from congenital heart disease data collected from thirteen clinical centers in China, with the initiative to analyze and compare clinical pathways for first-surgery patients (patients undergoing surgery for the first time). 
The data includes 6780 patient observations with 250 variables, spanning from hospital admission and preoperative examination to diagnoses, surgery, discharge, and follow-up visits.
To implement our analysis, we extract the observations of patients who were diagnosed with only ASD and treated with either PIC or MISC. 
There are only 1.16\% missing values in these observations, so we simply impute the data with K-Nearest Neighbor (number of neighbors set to 10).
The final data includes 2850 observations, with 16 pre-treatment variables and five surgery-related variables, some of which are manually integrated with multiple original variables to make the variable set more concise. 
Pre-treatment variables include the information generated on admission, through the inquiry and diagnosis, and from pre-surgery medical check results. 
For instance, on admission, we have: i) two cumulative scores on disease history \textit{chdhis} and \textit{othdhis}. 
If a patient's \textit{chdhis}$=3$, it means that the patient has three previous cardiovascular conditions, ii) basic measurements with \textit{sysbp} and \textit{diabp} representing systolic and diastolic blood pressure, respectively, and \textit{BMI} the Body Mass Index, and iii) demographic information like age and \textit{sex}. 

We give details of the ASD data in Appendix \ref{apx:ASD_data}, where Figure \ref{fig:realdata.des} shows the variable details and Table \ref{tab:des.stat.realdata1} and Table \ref{tab:des.stat.realdata2} show the descriptive statistics for the pre-treatment variables and the surgery-related variables (and outcome HFD1Y), respectively.

Before we apply CRL to this observational data, we first need to pre-process the data using the propensity score matching method \cite{zhao2021propensity}, which has proven to be very useful in estimating treatment effects using observational data \cite{Psmcomparison}.
This process allows us to match the data samples from both treatment groups to achieve pseudo-randomization of treatment assignment and to overcome the unsatisfactory overlap between the two groups, thus ensuring the satisfaction of the unconfoundedness assumption and overlap assumption.
Additionally, this process models propensity scores that can be used in the D-learning step.
Details of the matching process can be found in Appendix \ref{apx:psm}, after which 26 observations are eliminated from the data and variable \textit{atsize0} is replaced by its square root, $sq.atsize0$, for better covariate balance.
We then use the matched data from this process to fit all models for performance evaluation. 

\subsubsection{Performance metrics for real-world observational data}
\label{realworld.metrics}
\begin{figure}
	\centering
	\includegraphics[scale=.9]{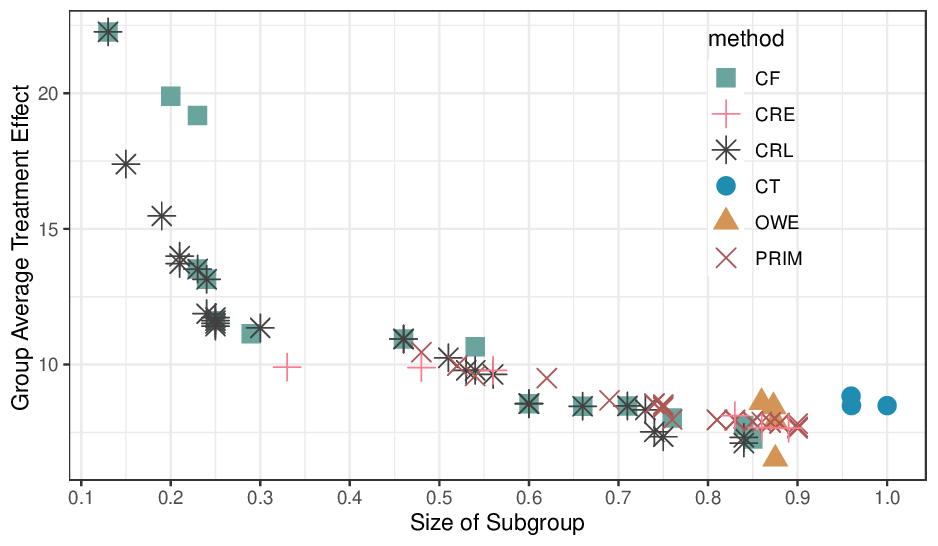} 
	\caption{Treatment efficient frontiers of all methods fit for 100 repetitions.}
	\label{fig:tef}	
\end{figure}

When estimating the treatment effect using real-world data, the ground truth is never known. 
Thus, we can no longer use the performance metrics defined for simulated data. Instead, we have to do so in some indirect ways based on what we know, like the following: 

i)  ITE-based prediction accuracy.
It is reasonable to evaluate the potential accuracy of ITE estimation based on the prediction accuracy of known observed outcomes $Y$. 
Consider $Y(-1) = f(X) + \epsilon$ which merely reflects the baseline effect and $Y{(1)} = Y{(-1)}+ \tau(X)$.
Then for the same outcome prediction model $\hat{Y} = g(\hat\tau(X),x)$ using entirely the same variables $x$ but different ITE estimates $\hat\tau(X)$, the difference in prediction accuracy, say $MSE(g(\hat\tau(X))$ mainly comes from the difference in $\hat\tau(X)$. Hence, in general, the method that predicts the outcome better (lower MSE) estimates \taux better. 

ii) Empirical expected outcome (EEO). This metric is defined in \cite{qi2018d} as:
$$ \hat{V}(d) = \frac{\mathbb{E}_N [Y1(A = d(X))/\pi (A,X)] }
{\mathbb{E}_N [1(A = d(X))/\pi (A,X)]},
$$ 
where $\mathbb{E}_N$ denotes empirical average and $d(X)$ denotes a specific personalized treatment strategy that maps a given $X$ to a certain treatment. 
This metric, in essence, measures the mean magnitude of the observed outcomes $Y$ that a treatment strategy achieves when its treatment recommendations align with the treatments actually received by the samples. 
The higher this expected outcome, the better the treatment strategy is, and implicitly, the more accurate our ITE estimation is.

iii) Mean-squared prediction error (MSPE) of transformed outcome.
Proposed in \cite{hitsch2024heterogeneous} as $\operatorname{MSPE} = \mathbb{E}[(Y_i^* - \hat{\tau}(X_i))^2]$. 
The transformed outcome can be calculated from the observed outcome $Y_i$ and propensity score $\pi(1,X)$:
\begin{equation*}
    Y_{i}^{*}=\frac{A_i + 1}{2} \cdot \frac{Y_{i}}{\pi(1,X_i)} - \frac{1 - A_i}{2} \cdot \frac{Y_{i}}{1-\pi(1,X_i)}.
\end{equation*}
$Y_{i}^{*}$ is a well-established unbiased estimate of $\tau(X_i)$ under the unconfoundedness assumption, i.e., $\mathbb{E}[Y_i^* | X_i] = \tau(X_i)$. Consequently, the MSPE serves as a proxy for the unobservable mean-squared error $\mathbb{E}[(\tau(X_i) - \hat{\tau}(X_i))^2]$. 
Therefore, an estimator with a lower MSPE is considered to provide a more accurate estimate of the true treatment effect.

iv) Treatment efficient frontier, proposed in \cite{wang2021causal}, is a graphic comparison of subgroups obtained from different methods, given that the identified subgroups differ in group size (proportion of samples) and CATE.
Though not always the case, a small group usually has a higher CATE than a bigger one, so it is hard to tell which subgroup is better. 
To balance the trade-off between subgroup size and subgroup CATE, the authors define the concept of treatment efficient frontier, which is formed with all dominant groups (subgroup that has no other subgroup surpassing it in both size and CATE) a certain method identifies. 
This frontier is then mapped to the two-dimensional coordinate system with the $x$-axis being the size and  $y$-axis the CATE estimates of the dominant groups. 
With this graph, we can see how quickly the estimated treatment effect magnitude decays as the subgroup covers more samples and easily spot the method that has the highest frontier since it finds subgroups with larger sizes and CATE.  
See Figure \ref{fig:tef} for example. 

\begin{table}[htbp]
	\centering
	\caption{Mean squared error (MSE) of outcome prediction and empirical expected outcome (EEO) averaged over 100 repetitions of ASD data applying CRL and other baseline models. SD is the standard deviation of MSE and EEO. }
	\begin{tabular}{lll}
		\hline
		& Mean MSE (SD)& Mean EEO (SD)\\
		\hline
		CRL& 21.608
        (12.840)
        & 360.514
        (0.763)\\
        CF& 21.619
        (12.844)& 360.732
        (0.338)\\	
         CT& 22.513 (13.817)& 359.729 (0.281)\\
         OWE& 21.582
        (12.831)& 360.569
        (1.042)\\
         PRIM& 21.622
        (14.541)&359.691
        (0.323)\\
         CRE& 21.112
        (14.001)&359.062
        (1.154)\\
        \hline
    \end{tabular}
\label{tab:realresult}
\end{table}

Considering the above performance metrics, especially the ITE-based prediction accuracy, we implement all methods: We first split the ASD data into three parts: 40\% (estimation set) for ITE estimation with all methods; 40\% (prediction set) for outcome prediction using the least squares model (or any other proper prediction models) fed with all the same variables except the different ITE estimated by different methods in the previous step. 
The remaining 20\% data (test set) is used to report the first two performance metrics, \ie $MSE(g(\hat\tau(X))$ and  EEO. 
We repeat the above procedure 100 times with randomly re-sampled ASD data.

To compute the MSPE, we employ a bootstrapped sampling approach: we split our ASD data into a training set (70\% samples) for fitting the estimation model and a test set (30\% samples) for reporting MSPE. 
We then draw 1,000 bootstrapped samples with replacement from each dataset, ensuring no data sample is contained in both the training and test sets. 
The MSPE values reported represent the average across these 1000 bootstrap samples.
In both the application of CRL to compute the above three metrics, we tune the relevant parameters using the same approach as in the simulation part. See Appendix \ref{apx:para_tune} for details.

To evaluate the treatment efficient frontier,  we use the entire data to fit all models for each of the resampled data, since CRL generally needs more observations to discover potential subgroups.
For each method to be compared, we pool the subgroups identified from all 100 repetitions together and find its dominant subgroups, which form the ultimate frontier.

\subsubsection{Model results and interpretation}
\begin{table}[t]
	\centering
	\caption{Mean-squared prediction error (MSPE) of transformed outcome with standard error (SE) and standard deviation (SD) across the 1000 bootstrap test samples of ASD data applying CRL and other baseline models. }
	\begin{tabular}{llll}
		\hline
		& Mean MSPE&SE &SD\\
		\hline
		CRL& 81775.58
       & 270.61& 8557.44\\
        CF& 1184278       & 12962.8&409919.7\\	
         CT& 1183005& 12975.51&410321.8\\
         OWE& 1166325& 13303.69&420699.7\\
         PRIM& 1187621& 13147.51&415760.9\\
         CRE& 1184111& 12961.91&409891.7\\
        \hline
    \end{tabular}
\label{tab:MSPE}
\end{table} 
We list the MSE of the predicted HFD1Y and EEO of all methods averaged over 100 repetitions in the first and second rows in Table \ref{tab:realresult}. 
It is obvious that all methods show similar accuracies of outcome prediction (MSE) and recommendation performance (EEO) on the ASD data. 
The summary statistics of MSPE are shown in \ref{tab:MSPE} where CRL demonstrates dramatically lower values compared to all baseline methods.
Figure \ref{fig:tef} shows the treatment efficient frontier of all methods. 
Obviously, compared with other methods (except CF), CRL identifies subgroups with a variety of sizes ranging from 0.1 to 0.85 and effect sizes ranging from 4 to 25. 
In contrast, methods like OWE and CT only identify big subgroups with relatively small subgroup CATE. 
The overlap between the CF and CRL frontiers is primarily driven by the CRL methodology itself: since CRL selects its rules from candidates generated by CF, its frontier inherently constitutes a refined, elite subset of the original CF rules, even after substantial filtering. Crucially, this demonstrates CRL's ability to achieve predictive performance comparable to CF while yielding a more parsimonious, interpretable, and clinically actionable set of subgroups.
The above results all demonstrate the effectiveness and superiority of CRL compared to other baseline methods on estimating HTE accurately.
Moreover, it provides us with an enhanced understanding of ITE of ASD treatments on HFD1Y through the informative rules we learned from the whole workflow.

For a concise presentation of what we learned from the ASD data with CRL, we list the eight rules that impact the estimated treatment effect the most in a randomly chosen rule set out of the 100 rule sets, with their corresponding CATE estimates and learned weights (See Table \ref{tab:rule2intpre}). 
The rules are displayed in decreasing order of the absolute value of the signed product of subgroup CATE multiplied by subgroup weight.
For example, the first rule means that when a patient with an abnormal ECG result, BMI$>$14.4, othchd$>$0, and is operated at a one-stop operating room, the patient's ITE of PIC against MISC will increase by 349.44 days. 
However, this estimate will then decrease by 323.34 if the patient's age is between five and nine. 
The final ITE estimate of a patient is the summation of all the products corresponding to the subgroups a patient belongs to, showing how a patient's ITE is affected by the combination of potential rules. 

These eight rules identify several important factors, such as age, BMI, electb, oroom, and NYHA, with meaningful cut-off values (e.g., 14.44 for BMI), influencing the treatment effect of PIC versus MISC on HFD1Y. 
Age and BMI are the two most involved factors in the rules, indicating they play indispensable roles in the way HFD1Y is affected by PIC and MISC. 
This is consistent with \cite{hughes2002prospective} and \cite{qi2020open} stating that age and weight-related factors like BMI are particularly important in surgery decisions and the prognosis of ASD, such as length of hospital stay. 
Compared to MISC, PIC, in general, has the advantage of shorter hospital stay and is more suitable for non-underweight children or older children \cite{qi2020open}, which is reflected with the HFD1Y increase in the first (BMI$>$14.4) and second rule (age$>$8).
For children aged between five and nine (second rule), the decrease in HFD1Y or the increase in hospital stay may be due to a lower percentage of PIC-treated patients within the group compared to that of the remaining population (81\% versus  85\%). 
This reveals the preference for MISC over PIC in the young-age group, which has been a fact in past ASD treatment practice \cite{karamlou2008rush}.   
\begin{table}
    \centering
    \caption{Eight CRL subgroups that have the greatest impact on the estimates of treatment effect, displayed in descending order of the absolute value of the subgroup CATE multiplied by the subgroup weight.}
    \begin{tabular}{llll} 
    \hline
         Subgroup&  CATE&  Weight &  CATE\\
         &&&$\times$Weight\\
         \hline
	BMI$>$14.4 and electb=1 and oroom=1 and othchd$>$0 & 2.92 & 119.54 & 349.44 \\ 
		age$\le$9 and age$>$5 & 7.62 & -42.42& -323.34\\ 
		age$>$8 and NYHA$\le$1 & 5.98 & 50.70 & 303.11 \\ 
		age$>$7 and NYHA$\le$1 and sq.atsize0$\le$4.8 & 7.09 & -42.64& -302.33\\ 
		murmur$\le$2 and sysbp$>$100 & 7.79  & -31.14& -242.45\\ 
		age$>$5 and BMI$>$17.2 and electb=1 & 6.09 & -38.38& -233.79\\ 
		age$>$15 and BMI$>$13.7 & 6.85 & -33.52& -229.63\\ 
		diabp$>$59 and electb=1 and oroom=0 & 6.80 & 29.20 & 198.66 \\ 
  \hline
    \end{tabular}
    \label{tab:rule2intpre}
\end{table}
In addition, these rules identify multiple cut-off values for the same factors, such as age with thresholds at 8 (third rule) and 15 (seventh rule), together with NYHA and sq.atsize0, revealing potential complex interactions between these factors. 
Although there is little evidence of these interactions, we believe they may stimulate further in-depth exploration by researchers and yield promising findings for actionable suggestions in clinical practice.
Specifically, these rules can be used to guide the stratification design of RCTs, which may improve the allocation of limited experimental resources. 
Take the third rule age$>$8 and NYHA$\le$1 in Table \ref{tab:rule2intpre} for example, researchers can first collect available external datasets to check if this rule still discriminates the corresponding subpopulation and the rest population on treatment effects.
If so, experimentalists can design targeted RCTs by stratifying subjects based on age and NYHA (and their potential cut-off values 8 and 1).
Once the RCT results demonstrate significant treatment effect heterogeneity across different groups, warnings like "this patient may have longer HFD1Y when treated with PIC than MISC due to age$>$8 and NYHA$\le$1!" can be incorporated into the decision support system and be triggered when a physician decides to perform PIC surgery on a patient.
Moreover, such rules can also be integrated as scoring items into clinical scoring systems to assess patient prognostic outcomes.

\begin{table}[h]
     \centering
     \caption{The dominant subgroups identified by CRE, displayed in descending order of subgroup CATE.}
     \begin{tabular}{lll}
     \hline
          Subgroup& CATE &Size\\
          \hline
 othchd$>$0 and sq.atsize0$>$0 and electb$\le$0& 9.91 &0.33\\
 	
atrial$>$2 and electb$\le$0&9.89 &0.48\\
 atrial$\le$2 and electb$\le$0&9.79 &0.56\\
 mesh$\le$0 and BMI$\le$22&8.12 &0.83\\
 atrial=2& 7.68 &0.86\\
 chdhis$\le$2 and sq.atsize0$>$0 and BMI$\le$22
&7.66 &0.89\\
\hline
     \end{tabular}
     \label{tab:CRE_rules}
 \end{table}

We also list the details of the dominant subgroups of CRE (See Table \ref{tab:CRE_rules}) and PRIM (eight top rules by subgroup CATE, see Table \ref{tab:PRIM_rules}) shown in Figure \ref{fig:tef} for a better comparison on interpretability.  
Apparently, both CRE and PRIM identify relatively homogeneous rules that entail limited information.
We find that for the same variable (except murmur), its cutoff value is almost the same in all the rules involved, such as 2 for atrial, NYHA, and CHDHI across all rules.
Many rules share similar and even identical variables, such as atrial$>$2 and electb$\le$0, atrial$\le$2, and electb$\le$0 (the second and third rules) in Table \ref{tab:CRE_rules} and the first two rules in Table \ref{tab:PRIM_rules},
which further exacerbates this rule of homogenization.
As a result, the subgroups identified are highly overlapping, hence the estimates of their treatment effects are also quite close.
In contrast, CRL can explore a more diverse combination of variables to uncover potential relationships in the ASD data. 
When different variables interact with each other, CRL identifies different cut-off values for each variable, which is more in line with the actual situation of complex disease, therefore providing much more interpretation and insights to healthcare practitioners.

\section{Conclusions and future work}\label{conclusion} 
This study introduces a rule-based framework, CRL, aimed at enhancing our understanding of HTE estimation for complex disease treatment in clinical scenarios. 
CRL leverages the causal forest to discover a pool of potential rules for ITE estimation (rule discovery) and then utilizes the D-learning method to filter out non-informative rules, imposing sparsity on the rules (rule selection).  
By deconstructing the estimated ITE through a weighted combination of the informative rules selected by D-learning, CRL explores HTE with a novel perspective that connects group-level and individual-level treatment effects. 
The understandable rules, their corresponding CATE estimates, weights in combination, and the comprehensive procedure for evaluating each causal rule from multiple perspectives (rule analysis) together give insights into the treatment of a complicated disease.
The remaining rules that survive after the rule analysis step may reveal potential meaningful interactions between involved variables, paving the way for further investigation and validation. 
\begin{table}
\centering
\caption{The eight top dominant subgroups identified by PRIM, displayed in descending order of subgroup CATE.}
\vskip 0.3cm
\begin{tabular}{lcc}
\hline
Subgroup& CATE& Size\\
\hline
NYHA$\le$2 and othdhis$\le$0 and occnum$\le$1 & & \\
 and chdhis$\le$2 and atrial$\le$2 and mesh$\le$0& 10.46&0.48\\
  and othchd$\le$5 and surgesitua$\le$1 and murmur$\ge$3& &\\
 & &\\
NYHA$\le$2 and chdhis$\ge$1 and chdhis$\le$2 and &&\\
 surgesitua$\le$1 and occnum$\le$1 and atrial$\le$2& 9.96&0.52\\
  and othdhis$\le$0 and mesh$\le$0 and oroom$\le$0 and murmur$\ge$2& &\\
 & &\\

 chdhis$\le$2 and NYHA$\le$2 and murmur$\ge$3&9.59&0.54\\
 & &\\
 NYHA$\le$2 and occnum$\le$1 and chdhis$\le$2 and othdhis$\le$0 and &9.50&0.62\\
 mesh$\le$0 and surgesitua$\le$1 and murmur$\ge$2& &\\
 & &\\
 oroom$\le$0 and NYHA$\le$2 and chdhis$\ge$1 and & &\\
 occnum$\le$1 and chdhis$\le$2 and othchd$\le$5& 8.68&0.69\\
 and mesh$\le$0 and atrial$\le$2 and othdhis$\le$0& &\\
 & &\\
 oroom$\le$0 and occnum$\le$1 and NYHA$\le$2 and atrial$\le$2&8.56&0.74\\

 NYHA$\le$2 and oroom$\le$0 and atrial$\le$2& 8.55&0.74\\
 NYHA$\le$2 and chdhis$\le$2 and chdhis$\ge$1 and oroom$\le$0& 8.53&0.75\\
 NYHA$\le$2 and surgesitua$\le$1 and oroom$\le$0& 8.48&0.75\\
 surgesitua$\le$1 and oroom$\le$0& 8.42&0.75\\
\hline
\end{tabular}
\label{tab:PRIM_rules}
\end{table} 
This research has several methodological limitations that could be addressed in future studies. 
First of all, CRL requires the outcome of interest to be continuous and prefers higher values. 
Treatment in our study is also limited to the binary case. 
These limits hinder the wide application of the method in healthcare practice.
Secondly, our assumption of ITE is based on a linear combination of potential rules for the simplicity of demonstration, while the non-linear case is not discussed and evaluated.
Thirdly, in the evaluation of CRL from treatment recommendation, we simply recommend treatments based on the sign of the estimated ITE and do not consider the cost-sensitive scenarios where treatment resources are constrained.
Future work may introduce more economic factors and explore how to better estimate treatment effects and recommend treatments in complex scenarios.
Lastly, while the current staged architecture of CRL achieves superior interpretability and clinical practicality, it may result in lower efficiency compared to a fully integrated framework. We will explore the possibility of a more unified approach that maintains interpretability while improving operational efficiency in future work.

\section*{Conflicts of Interest}
The authors declare no conflicts of interest.

\section*{Data Availability Statement}
The real-world data that support the findings of this study contain sensitive patient information and, therefore, cannot be made publicly available. However, these data are available from the corresponding author upon reasonable request, subject to appropriate confidentiality agreements.

\bibliographystyle{tfs}
\bibliography{CRL_abb.bib}

\newpage
\appendix
\section{The PDR framework}
\label{apx:PDR}
\begin{figure}[h]
\centering
\includegraphics[scale=.1]{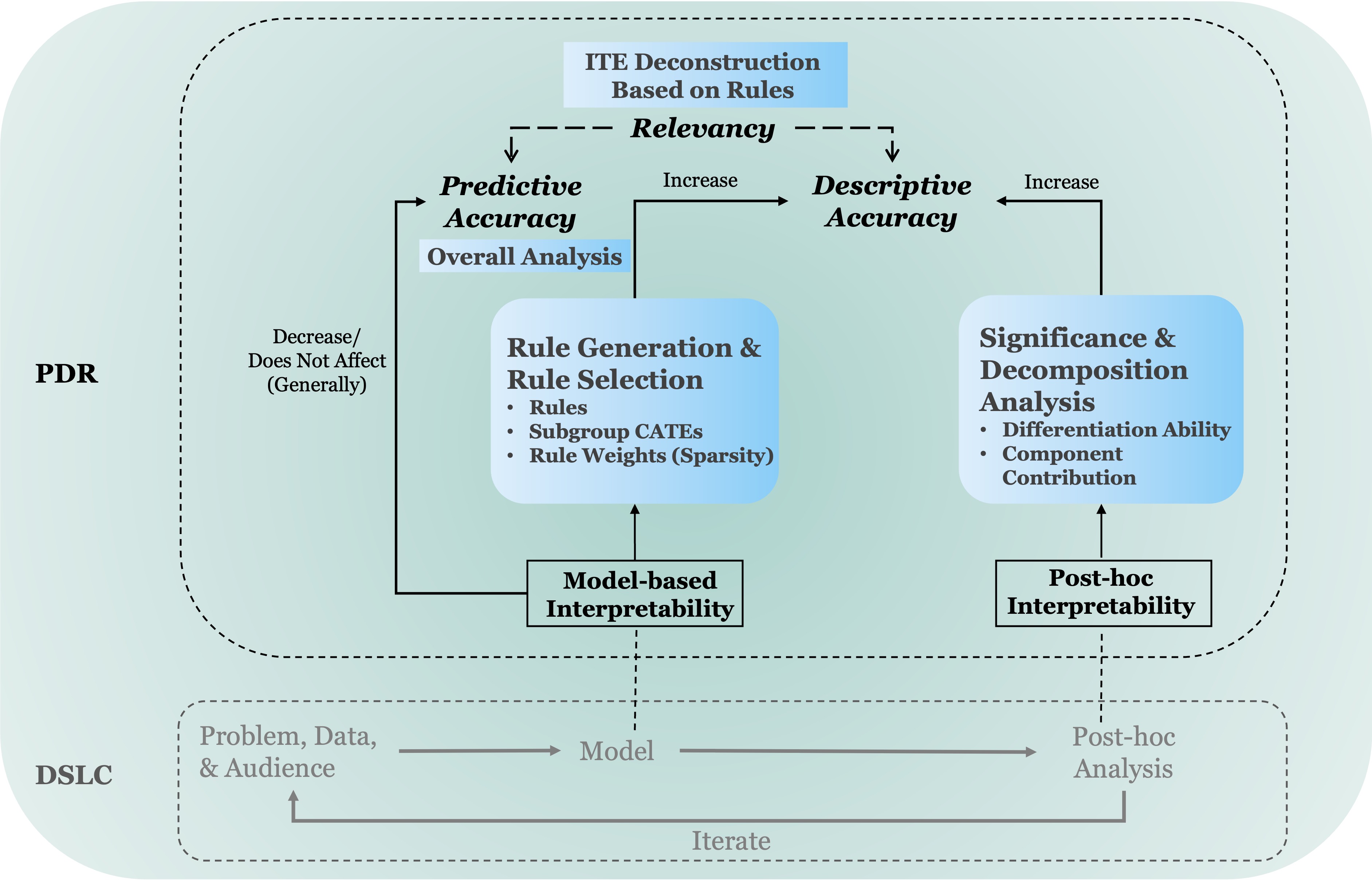}
\caption{Overview of the PDR framework of interpretability in different stages where interpretation matters in the data-science life cycle \cite{yubinPDR2019}. 
How CRL enables PDR interpretability is also shown and highlighted in blue blocks.}
\label{fig:dslc}
\end{figure}

The PDR framework \cite{yubinPDR2019} provides comprehensive guidance and a rich vocabulary for interpretability within the data science life cycle (DSLC) in relevant research \cite{wu2025newcausalrulelearning}. 
As shown in the lower part (gray texts) of Figure \ref{fig:dslc}, in a typical DSLC, we begin with a specific domain problem for a specific audience and certain data collected to study the problem. 
Then, in the modeling part, we explore the data and choose certain methods to fit it. After that, we analyze what the model has learned in the post-hoc stage.  
Figure \ref{fig:dslc} also presents the critical concepts in the PDR framework and their relationships in different stages of DSLC where interpretation matters. 
The framework consists of three desiderata of interpretations, namely predictive accuracy, descriptive accuracy, and relevancy (italic, bold black texts in the upper part). 
Predictive accuracy evaluates how well the chosen model fits the data. Descriptive accuracy is the degree to which an interpretation objectively captures the relationships learned by the model. 
Relevancy measures how much insight an interpretation provides into the research question for a particular audience. 
Interpretation is also classified into model-based and post-hoc interpretability (framed bold black texts). 
The first interpretability involves using a simpler model to fit the data during the modeling phase to increase descriptive accuracy, in which predictive accuracy generally decreases or remains unchanged.
The second one involves using various methods to extract information from a trained model to enhance descriptive accuracy. 
When choosing a model, there is often a trade-off between both accuracies (dashed line with a two-way arrow), \ie whether to use a black box model with higher predictive accuracy or a simpler model with higher descriptive accuracy?
The answer usually lies in relevancy determined by the problem's specific context and its audience.

\section{Three assumptions of Neyman--Rubin potential outcome framework}
\label{apx:rubin.assumps}
Below, we outline the three basic assumptions of the potential outcome framework:
\vskip 0.1cm
\textbf{1. Stable Unit Treatment Value Assumption (SUVTA):} The potential outcomes of any instance do not vary with the treatments assigned to other instances and there is only one version of each treatment.
\vskip 0.1cm
\textbf{2. Strong Ignorability/Unconfoundedness:}
Treatment assignment is independent of the potential outcomes conditional on the observed covariates, \ie $A \perp \{Y(1),Y(-1)\}|X = x$.  This assumption implies there are no other unmeasured confounding factors, and treatment assignment can be considered as random given the covariates $X$.

\cite{rosenbaum1983central} gives an equivalent assumption conditioned on the propensity score, \ie $A \perp \{Y(1),Y(-1)\}|\pi(a,X)$, which is especially useful in high-dimension data settings where it is difficult to match samples with the original covariates. Our method is based on this assumption.
\vskip 0.1cm
\textbf{3. Positivity/Overlap:}
$0<Pr(A = 1| X = x)<1$, \ie any instance has a positive probability of receiving either treatment given $X = x$. This assumption ensures the existence of samples for both treatment groups so that it is possible to estimate the treatment effect.

RCT setting is usually considered to satisfy these assumptions automatically due to its deliberate design  (especially the randomization on treatment assignment).
However, observational data have to be pre-processed to make their treatment allocation pseudo-random to satisfy the (Conditional) Unconfoundedness Assumption and to make their sample sizes more balanced between groups to achieve an acceptable or better overlap.

\section{Proof of Theorems and Lemmas }
\label{apx: theory}
\subsection{Proof of Theorem \ref{thm}}
The proof for Theorem \ref{thm} can be easily derived from Lemmas \ref{lem:1} and \ref{lem:2} given below.
\subsection{Additional Lemmas}
To prove Theorem \ref{thm}, we need two lemmas.
\begin{lemma}\label{lem:1}
    For any approximation $\hat{\tau}(\mathbf{X})$  of $f_0(\mathbf{X})$, the value difference between the $d^*(\mathbf{X})$ (induced by $f_0(\mathbf{X})$) and the estimated $\hat{d}_N$ (induced by $\hat{\tau}(\mathbf{X})$) can be bounded as 
    \begin{equation}\|V(d^*)-V(\hat{d}_N)\|_2\leq(\mathbf{E}\|f_0(\mathbf{X})-\hat{\tau}(\mathbf{X})\|_2^2)^{\frac{1}{2}}.\end{equation}
Furthermore, if we assume gMC holds, the upper bound can be optimized as 
    \begin{equation}\|V(d^*)-V(\hat{d}_N)\|_2\leq C'(\mathbf{E}\|f_0(\mathbf{X})-\hat{\tau}(\mathbf{X})\|_2^2)^{\frac{1+\alpha}{2+\alpha}}.\end{equation}
\end{lemma}

\begin{lemma}\label{lem:2}
    Assume assumptions (1)-(6) hold. For $t>0$, let the tuning parameter $\lambda$ be
    \begin{equation}\lambda=16\sqrt{2}t^2\sqrt{\frac{\log^2(2M)}{N}}.\end{equation}
    Then for the constant $C_1$ which depends on the constants $a, \rho$ and $\sigma^2$ and $t$ sufficiently large, with the probability at least $1 - \frac{C_1}{t^2}$, we have
$$\frac{2\|\tilde{\mathbf{X}}_N\hat{{\beta}} - f_0\|_2^2}{N} + 3\lambda\|\hat{{\beta}} - {\beta}^*\|_1 \leq 6\|\tilde{\mathbf{X}}_N{\beta}^* - f_0\|_2^2/N + \frac{24\lambda^2s_*}{\phi_*^2}.$$
\end{lemma}
The proof of the first lemma can be found in \cite{qi2018d}. 
We give the proof of the Lemma \ref{lem:2} below.

\subsection{Proof of Lemma \ref{lem:2}}
Before proof of Lemma \ref{lem:2}, we have 
\begin{lemma}[Nemirovski moment inequality]\label{lem:3}
    For $m\geq 1$ and $p\geq e^{m-1}$, we have the following inequality
$$\mathbf{E} \max_{1 \leq j \leq M} |\sum_{i=1}^{n}\left(\gamma_{j}\left(Z_{i}\right)-\mathbf{E} \gamma_{j}\left(Z_{i}\right)\right)^{m}| \leq[8 \log (2 M)]^{\frac{m}{2}} \mathbf{E}\left[\max _{1 \leq j \leq M} \sum_{i=1}^{n} \gamma_{j}^{2}\left(Z_{i}\right)\right]^{\frac{m}{2}} .$$
\end{lemma}
\begin{proof}[Proof of Lemma \ref{lem:2}.]
With some algebra derivations, we have
\begin{equation}\|\tilde{\mathbf{X}}\hat{\beta}-f_0\|_2^2/N+\lambda\|\hat{\beta}\|_1\leq2\epsilon^\top \tilde{\mathbf{X}}(\hat{\beta}-\beta^*)/N+\lambda\|\beta^*\|_1+\frac{\|\tilde{\mathbf{X}}\beta^*-f_0\|_2^2}{N}.\end{equation}
Then for the first term on the right hand, by the H\"{o}lder's inequality we can have
\begin{align*}
    2|\epsilon^\top \tilde{\mathbf{X}}(\hat{\beta}-\beta^*)/N|\leq(\max_{1\leq j\leq M}2|\epsilon^\top \tilde{X}^{(j)}|/N)\|\hat{\beta}-\beta^*\|_1.
\end{align*}
Let $\lambda_{0}=16\sqrt{2}t^{2}\sqrt{\frac{\log^{2}(2M)}{N}}$ and define a set
$$\Lambda:=\{2\max_{1\leq j\leq M}|\epsilon^{\top}\tilde{X}^{(j)}|/N\leq\lambda_{0}\}.$$

Through Lemma \ref{lem:3} with $m=2$, we have 
\begin{align*}
    P(\Lambda^{c})&=P(\{2\max_{1\leq j\leq M}|\epsilon^{\top}\tilde{X}^{(j)}|/N\geq\lambda_{0}\})\\&\leq\frac{\mathbf{E}\max_{1\leq j\leq M}4|\epsilon^\top \tilde{X}^{(j)}|^2/N^2}{\lambda_0^2}\\&\leq\frac{4[8\log(2M)/N]\mathbf{E}[\max_{1\leq j\leq M}\sum_{i=1}^{n}\epsilon_{i}^{2}\tilde{X}_{ij}^{2}/N]}{16\times32t^{2}\frac{\log^{2}(2M)}{N}}\\&\leq\frac{a^{2}\sum_{i=1}^{N}\mathbf{E}\epsilon_{i}^{2}/N}{16t^{2}\log(2M)}=\frac{a^{2}(4\|\mathbf{X}\gamma_{0}\|_{2}^{2}/N+4\sigma^{2})}{16t^{2}\log(2M)}\\&\leq\frac{a^{2}(\rho\|\gamma_{0}\|_{2}^{2}+\sigma^{2})}{4t^{2}\log(2M)}\\
    &\leq \frac{C}{t^2}.
\end{align*}
Then with probability at least $1-\frac{C}{t^2}$, for large enough $t$, we have
\begin{align*}
    \|\tilde{\mathbf{X}}\hat{\beta}-f_{0}\|_{2}^{2}/N+\lambda\|\hat{\beta}\|_{1}
    &\leq(\max_{1\leq j
    \leq M}2|\epsilon^{\top}\tilde{X}^{(j)}|/N)\|\hat{\beta}-\beta^{*}\|_{1}+\lambda\|\beta^{*}\|_{1}+\frac{\|\tilde{\mathbf{X}}\beta^{*}-f_{0}\|_{2}^{2}}{N}\\
    &\leq\lambda_{0}\|\hat{\beta}-\beta^{*}\|_{1}+\lambda\|\beta^{*}\|_{1}+\frac{\|\tilde{\mathbf{X}}\beta^{*}-f_{0}\|_{2}^{2}}{N}.
\end{align*}
Following the same procedure of Lemma 4 in \cite{qi2018d}, we have the following conclusion. That is,
    on the set $\Lambda$, with $\lambda\geq2\lambda_0$, 
    \begin{equation*}
        4\|\tilde{\mathbf{X}}\hat{\beta}-f_{0}\|_2^2/N+3\lambda\|\hat{\beta}_{S_*^c}\|_1\leq5\lambda\|\hat{\beta}_{S_*}-\beta_{S_*}^*\|_1+4\|\tilde{\mathbf{X}}\beta^*-f_0\|_2^2/N,
    \end{equation*}
    where $S_{*}=\{j:\beta_{j}^{*}\neq0\}$.
Therefore, 
\begin{align*}
    &\quad 4\|\tilde{\mathbf{X}}\hat{\beta}-f_{0}\|_{2}^{2}/N+3\lambda\|\hat{\beta}-\beta^{*}\|_{1}\\&=4\|\tilde{\mathbf{X}}(\hat{\beta}-\beta^{*})\|_{2}^{2}/N+3\lambda\|\hat{\beta}_{S_{*}}-\beta_{S_{*}}^{*}\|_{1}+3\lambda\|\hat{\beta}_{S_{*}^{c}}\|_{1}\\&\leq12\lambda\|\hat{\beta}_{S_{*}}-\beta_{S_{*}}^{*}\|_{1}\leq12\sqrt{s_{*}}\lambda\|\tilde{\mathbf{X}}(\hat{\beta}-\beta^{*})_{S_{*}}\|_{2}/(\sqrt{N}\phi_{*})\\&\leq\sqrt{s_{*}}\|\tilde{\mathbf{X}}\hat{\beta}-f_{0}+f_{0}-\tilde{\mathbf{X}}\beta^{*}\|_{2}/(\sqrt{N}\phi_{*})\\
    &\leq12\sqrt{s_{*}}\|\tilde{\mathbf{X}}\hat{\beta}-f_{0}\|_{2}/(\sqrt{N}\phi_{*})+12\sqrt{s_{*}}\|\tilde{\mathbf{X}}\beta_{*}-f_{0}\|_{2}/(\sqrt{N}\phi_{*})\\&\leq6\|\tilde{\mathbf{X}}\beta^{*}-f_{0}\|_{2}^{2}/N+\frac{24\lambda^{2}s_{*}}{\phi_{*}^{2}}+2\|\tilde{\mathbf{X}}\hat{\beta}-f_{0}\|_{2}.
\end{align*}
Therefore, 
$$\frac{2\|\tilde{\mathbf{X}}\hat{{\beta}} - f_0\|_2^2}{N} + 3\lambda\|\hat{{\beta}} - {\beta}^*\|_1 \leq 6\|\tilde{\mathbf{X}}{\beta}^* - f_0\|_2^2/N + \frac{24\lambda^2s_*}{\phi_*^2}.$$
\end{proof}

\section{Parameters tuning for CRL and baselines}
\label{apx:para_tune}
\setcounter{figure}{0}  
\setcounter{table}{0}
This section outlines several parameters that require tuning when fitting CRL and other baseline models.
All methods, including CRL and the baselines, are tuned using 10-fold cross-validation. 
The performance criteria are the mean squared error (MSE) of estimators and ITE-based prediction accuracy for the simulated data and the real-world ASD data, respectively. 

\subsection{Parameters tuning for CRL}
\begin{table}
    \centering
\caption{Summary of parameters tuning for CRL (\minnodesize not included).}
\label{tab:parameters}
    \begin{tabular}{lll}
    \hline
        \textbf{Parameters} & \textbf{Tuning Range} & \textbf{Final value chosen}\\
        \hline
         \textit{num.trees}& \{10, 50, 100,200,300,400,&50\\
        & 500,1000,1500,2000\}& \\
        \textit{mtry}& &\\
 \multicolumn{1}{r}{Simulated data}& \{3,4,\dots,8,9\}&9\\
 \multicolumn{1}{r}{ASD}& \{4,5,\dots,15,16\}& 16\\
        $\lambda$  & 100 values in [0.0004,4] & The one that gives the \\
        & with an interval of 0.04& minimum mean error in  \\
	& & 10-fold cross-validation\\
    \hline
    \end{tabular}
\end{table}

\begin{table}
    \centering
    \caption{Best value of \minnodesize for CRL under different simulated data settings, tuned with 20 values in [20,400] with an interval of 20.}
    \begin{tabular}{llll}
    \hline
         Number of Observations&  \multicolumn{3}{c}{Number of True Subgroups}\\
                 &1&3&5\\
         \hline
         2,500&70&50&80\\
         5,000&160&80&150\\
         10,000&300&150&190\\
         \hline
    \end{tabular}
    \label{tab:min.node.size4CRL}
\end{table}
In the rule discovery phase, we tune the parameters of CRL inherited from Causal Forest, such as \numtrees specifying the number of trees in the forest,  \minnodesize the minimum number of samples from both treatments in each tree leaf node used to estimate the treatment effect, and \mtry the number of variables tried for each split when fitting the forest. 
Other parameters of causal forest include \textit{honesty.fraction} and \textit{sample.fraction} specifying the sample allocation proportion in honest estimation and sample splitting, respectively. We set these two parameters to a fixed default value, \ie 0.5.
In the rule selection phase, the main parameter is $\lambda$ that controls the degree of penalty in Equation (\ref{def:dlearning}).

\begin{figure}
    \centering
    \includegraphics[scale = .7]{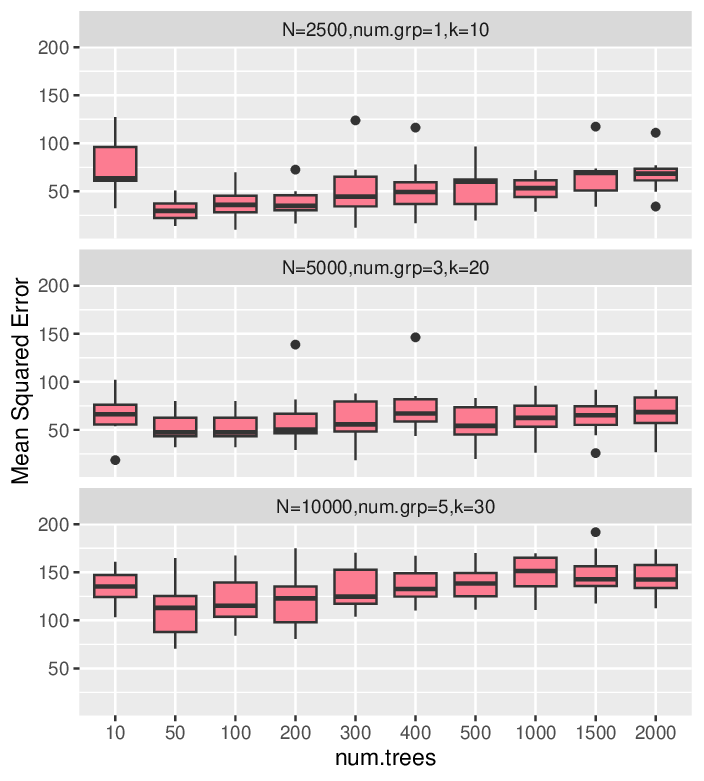} 
    \caption{Mean squared error of treatment effect applying CRL on 3 simulated datasets under different \textit{num.trees}.}
    \label{fig:num.trees_Tuning}
\end{figure}

Generally speaking, for both the simulated data and the ASD data in this study, we find that CRL's performance shows little improvement beyond 50 trees (when other parameters are set optimally).
We show Figure \ref{fig:num.trees_Tuning}, the results (mean squared error of CRL estimated effect) of tuning \textit{num.trees} with 3 simulated datasets for your reference.
Therefore, we set \textit{num.trees} to this fixed value to reduce computational costs and minimize the generation of irrelevant rules.
\mtry is tuned using integers from three to nine and four to sixteen, respectively, for simulated and ASD data, where nine and sixteen always correspond to the lowest MSE value.
For $\lambda$, the one that gives the minimum mean error is chosen from [0.0004,4]. Tuning details of all the above parameters are summarized in Table \ref{tab:parameters}.
\minnodesize appears to be the most sensitive parameter. 
Its performance is mainly affected by the number of observations and the number of pre-defined true (positive) subgroups. 
Details for simulated data are shown in Table \ref{tab:min.node.size4CRL}. For ASD data, the best value is 20.

\subsection{Parameters tuning for baselines}
We directly use the default parameter settings for OWE and PRIM since they have few parameters that require tuning.
Both CRL and CRE follow a rule-based workflow and generate rules using forest-based models.
Therefore, we set CRE parameters nearly the same as CRL, namely \numtrees to 50, and the mean depth of tree to 4. 
Other CRE parameters are set to default, such as the learning rate to 0.01.
 
Similar to CRL, the best \mtry for both CT and CF are still 9 for simulated data and 16 for the ASD data. 
\numtrees of CF is 50 for all data due to its insensitivity. 
For \minnodesize of CF, we find that 20 performs the best in all simulated data, and 40 for the ASD data. 
For CT applied to the ASD data, the optimal \minnodesize is 40
while for the simulated data, the value varies and is shown in Table \ref{tab:min.node.size4CT}.

\begin{table}
    \centering
    \caption{Best value of \minnodesize for CT under different simulated data settings, tuned with 20 values in [20,400] with an interval of 20.}
    \begin{tabular}{llll}
    \hline
         Number of Observations&  \multicolumn{3}{l}{Number of True Subgroups}\\
         &1&3&5\\
          \hline
         2,500&140&180&110\\
         5,000&220&280&190\\
         10,000&50&400&280\\
         \hline
    \end{tabular}
    \label{tab:min.node.size4CT}
\end{table}

\section{Robustness analyses of CRL with correlated covariates}
\label{app:correlated_data simulation}

\subsection{Generation of correlated data.}

We generate simulated data identical to the data used in our simulation study one but only introduced controlled correlations between specific variables: $r(x_1,x_5) = \gamma \times 0.45$, $r(x_3,x_4) = \gamma \times 0.34$, $r(x_1,x_5) = \gamma \times 0.1$, $r(x_1,x_5) = \gamma \times 0.2$. 
Here, $ \gamma \in \{0,1,2\}$ scales the correlation strength between variables, namely no correlation ($ \gamma =0  $), mild correlation ($ \gamma =1  $) and strong correlation ($ \gamma = 2  $) between variables. 
Since $\gamma=0$ corresponds to the original data used in our simulation, we only generate data for $\gamma=1,2$ for all possible combinations of num.obs, num.grp and k. 
Therefore, for each $\gamma$, we have 54 data sets, and each data set is resampled 100 times to fit CRL and other baseline methods. 
These datasets are used both for the overall performance evaluation of CRL and the robustness of p-value as well as its interpretation in the rule decomposition analysis.
\begin{table}
    \centering
    \caption{Details of three types of rules.}
    \begin{tabular}{cc}
    \hline
       Rule Type &  Rule \\
    \hline
    \multicolumn{2}{c}{Data Setting: $\{N=2500,num.grp=1,k=10\}$}\\
    \hline
           true &  $x_1=1  $ and $ x_2=1$ \\
           fake without correlation &  $x_1=1  $ and $ x_8>0$ \\
           fake with correlation &  $x_1=1  $ and $ x_5>0$ \\
           $r(x_1,x_5) = 0/0.45/0.9 $&  $x_1=0  $ and $ x_5\le0$ \\
    \hline
    \multicolumn{2}{c}{Data Setting: $\{N=5000,num.grp=3,k=20\}$ }\\
    \hline
           true &  $x_1=1  $ and $ x_1+x_2+x_3\ge1$ \\
           fake without correlation&  $x_1=0  $ and $ x_2=0  $ and $ x_8>0$ \\
           fake with correlation &  $x_1+x_2+x_3\ge3  $ and $ x_4>0$ \\
           $r(x_3,x_4) = 0/0.34/0.68 $&  $x_1+x_2+x_3=0  $ and $ x_5\le0$ \\
    \hline
    \multicolumn{2}{c}{Data Setting: $\{N=10000,num.grp=5,k=30\}$}\\
    \hline
             true & $x_1+x_2\ge1  $ and $ x_3>5$\\
             fake without correlation& $x_1+x_2=0  $ and $ x_8>0$\\
             fake with correlation& $x_1+x_2\ge1  $ and $ x_5>0$\\
             & $x_3>5  $ and $ x_4\le0$\\
    \hline
    \end{tabular}
    \label{tab:rule.types}
\end{table}
\subsection{Performance evaluation of CRL under correlation}
\begin{figure}
    \centering
    \includegraphics[width=1\linewidth]{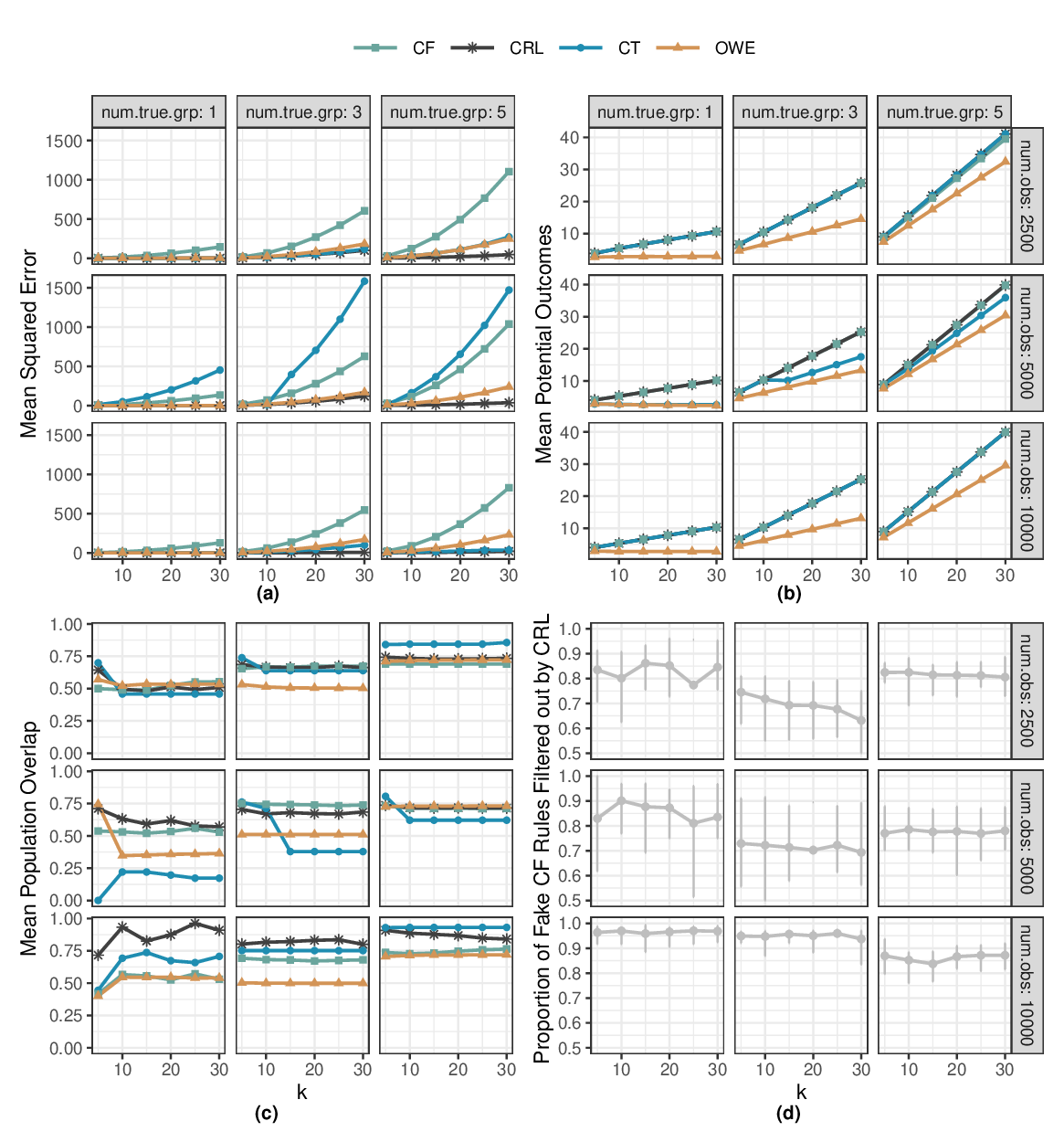} 
    \caption{Performance comparison of CRL (black asterisk) and baseline methods CT (blue circle), CF (green square), OWE (yellow triangle) applied on the correlated ($\gamma=1$) simulated data. 
    All performance metrics are averaged over 100 repetitions for each data set.
    (a) Mean squared error of treatment effect. 
 (b) Mean potential outcomes. (c) Mean population overlap. (d) Proportion of fake CF rules filtered out by CRL.}
    \label{fig:peformance_1}
\end{figure}

\begin{figure}
    \centering
    \includegraphics[width=1\linewidth]{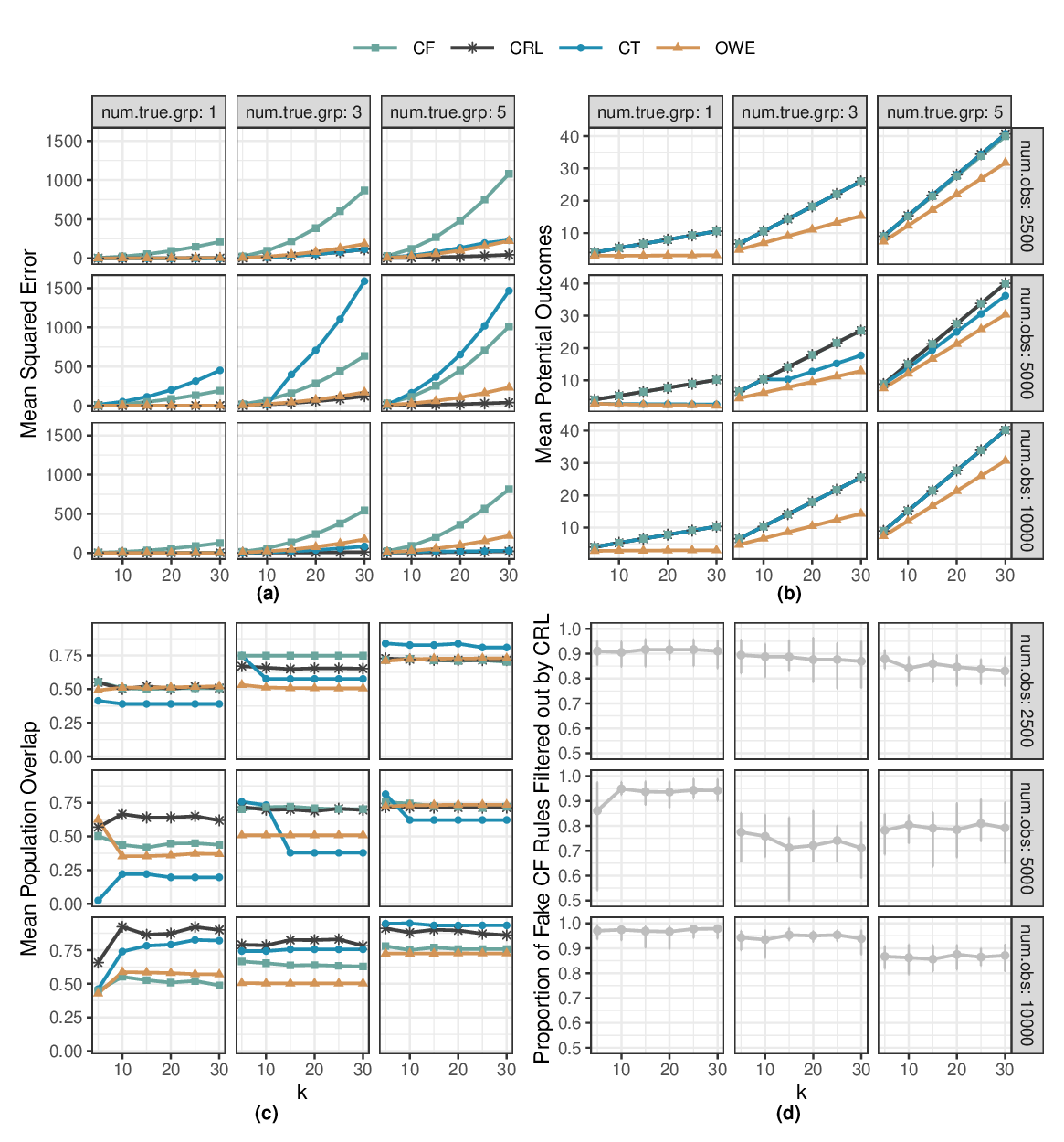} 
    \caption{Performance comparison of CRL (black asterisk) and baseline methods CT (blue circle), CF (green square), OWE (yellow triangle) applied on the correlated ($\gamma=2$) simulated data. 
    All performance metrics are averaged over 100 repetitions for each data set.
    (a) Mean squared error of treatment effect. 
 (b) Mean potential outcomes. (c) Mean population overlap. (d) Proportion of fake CF rules filtered out by CRL.}
    \label{fig:peformance_2}
\end{figure}

The results are shown in Figure \ref{fig:peformance_1} and \ref{fig:peformance_2} for $\gamma=1$ and $\gamma=2$ respectively, which indicates 
that variable correlation barely harms the performance of CRL, and that the superiority of CRL over other methods in uncorrelated cases remains.
Specifically, CRL keeps the lowest MSE (Figure \ref{fig:peformance_1}(a)  and \ref{fig:peformance_2}(a) ) and the highest MPO (Figure \ref{fig:peformance_1}(b)  and \ref{fig:peformance_2}(b)) together with CF in all settings for both $\gamma$.
Also, CRL has comparable performance to baseline methods with regard to PO (Figure \ref{fig:peformance_1}(c)  and \ref{fig:peformance_2}(c)) across most settings.
In addition, the proportion of fake CF rules filtered out by CRL (Figure \ref{fig:peformance_1}(d)  and \ref{fig:peformance_2}(d)) remains high, suggesting CRL's ability to distinguish true rules from fake ones, even when  there are redundant variables highly correlated with critical variables.

\subsection{Validation of rule decomposition interpretation}
\textbf{Three Types of Rules for Testing.} The variables generated in this section can be classified into three categories: set of critical variables which defines the true subgroup \ie $\{x_1,x_2,x_3\}$, set of redundant variables which are related to certain critical variables when $\gamma \ne 0$ \ie $\{x_4,x_5\}$ and the set of the rest variables. 
When a redundant variable, strongly correlated with a critical one, is removed, the resulting $p$-values might be influenced. 

To check if this issue matters, we create various rules by selecting variables from the above sets and check how the p-values of interest change in different combinations of uncorrelated/correlated variables. 
Specifically, we create three types of rules: true rules, fake rules without variable correlation, and fake rules with variable correlation. 
Details are shown in Table \ref{tab:rule.types}.
Note that for simplicity, we only test the results of the following three combinations of num.obs, num.grp, and k:  $\{N=2500,num.grp=1,k=10\}$ , $\{N=5000,num.grp=3,k=20\}$, $\{N=10000,num.grp=5,k=30\}$.  
For each data setting and a given $\gamma$, we only use 10 random datasets to obtain the results. 
Besides, the rules for the same type differ across data settings as the definition of true subgroups differs (See Figure \ref{fig:num.grp} for the definition of true subgroups). 
Additionally, due to the complex nature of rule definition and varying magnitude of positive treatment effect for settings with 3 and 5 true subgroups, the rules for testing are represented in a way to cover as many true populations as possible to avoid trivial comparison.
For example, we use $x_1=1  $ and $ x_1+x_2+x_3\ge1$ for $num.grp=3$, instead of testing multiple true combinations of $x_1,x_2,x_3$.

We then perform rule decomposition analysis for the rules shown above and compare the p-values of revised rules to the complete rule. 
For each revised rule, we compute $\Delta p = p_{revised}-p_{complete}$, \ie the difference of p-values between the revised rule and the complete rule. 
In terms of our decomposition analysis, a positive difference ($\Delta p>0$) means the component removed is critical to the complete rule; otherwise, it might be redundant. 
Therefore, a rule with a redundant variable (or negative $\Delta p$) will not survive in the rule decomposition analysis step and will be considered as not informative enough for further human validation.
Since each rule is analyzed on 10 resampled data, we count the number of positive and negative $\Delta p$ respectively and compare these numbers across $\gamma = 0,1,2$.

\begin{figure}
    \centering
    \includegraphics[width=.9\linewidth]{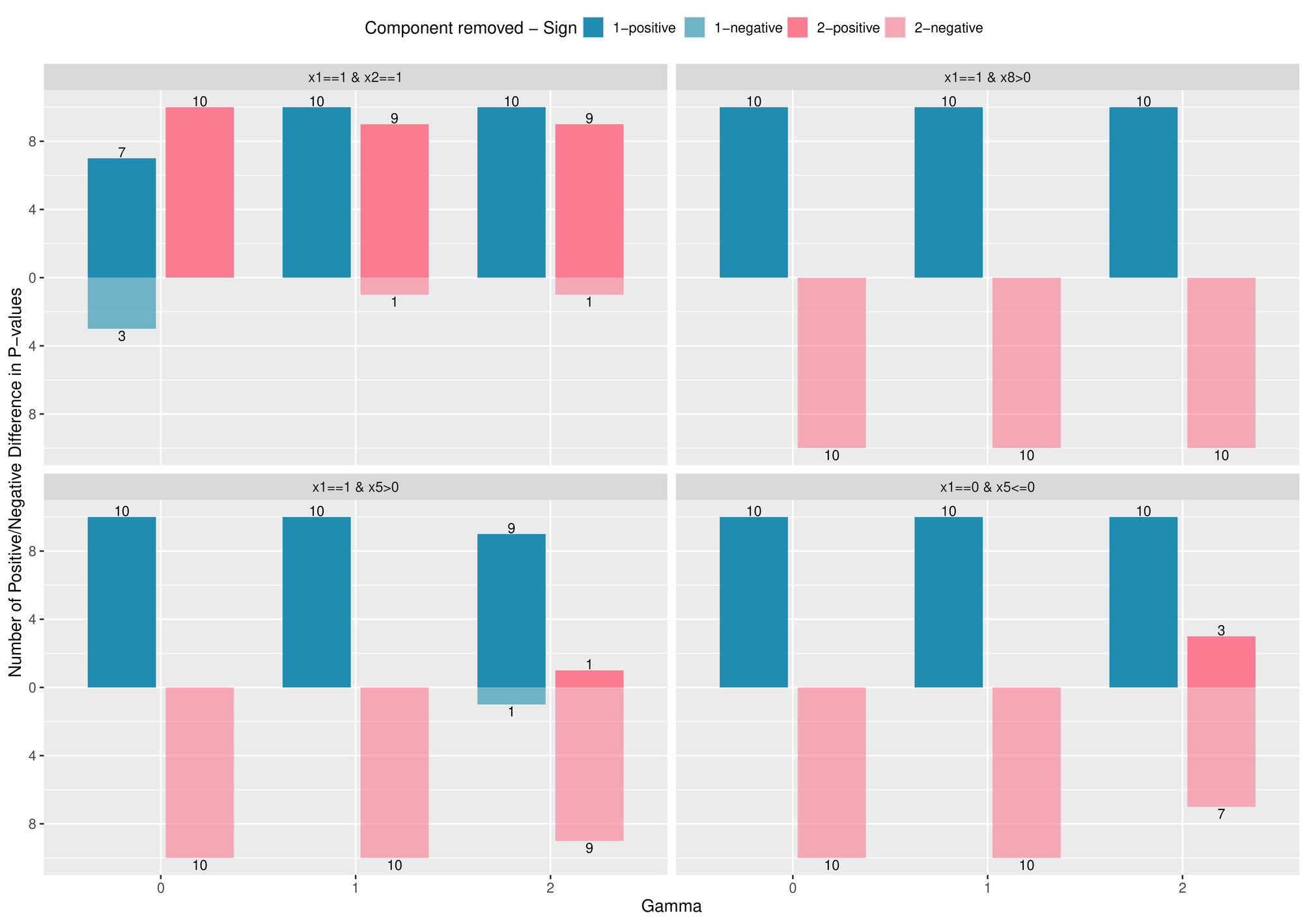}
    \caption{Results of rule decomposition analysis under different strengths of variable correlations ($\gamma=0,1,2$) for $\{N=2500,num.grp=1,k=10\}$.
    The pillars show the count of positive (above 0  vertical line) and negative (below 0 vertical line) $\Delta p$ corresponding to the first component removed (blue) and the second component (pink) removed.
}   
\label{fig:p-1grp}
\end{figure}

\begin{figure}
    \centering
    \includegraphics[width=.9\linewidth]{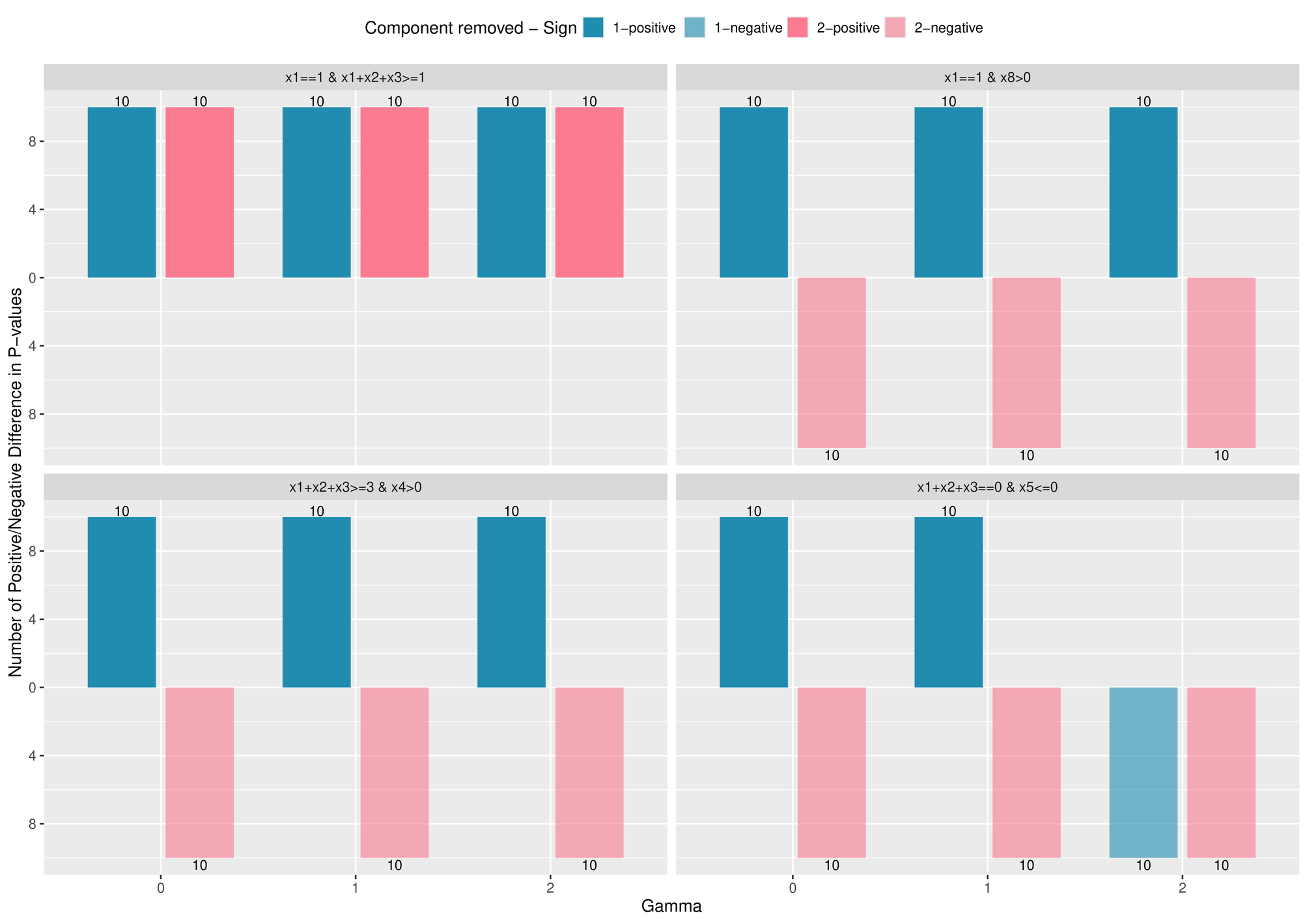}
    \caption{Results of rule decomposition analysis under different strengths of variable correlations ($\gamma=0,1,2$) for $\{N=5000,num.grp=3,k=20\}$.
    The pillars show the count of positive (above 0  vertical line) and negative (below 0 vertical line) $\Delta p$ corresponding to the first component removed (blue) and the second component (pink) removed.}
    \label{fig:p-3grp}
\end{figure}

\begin{figure}
    \centering
    \includegraphics[width=.9\linewidth]{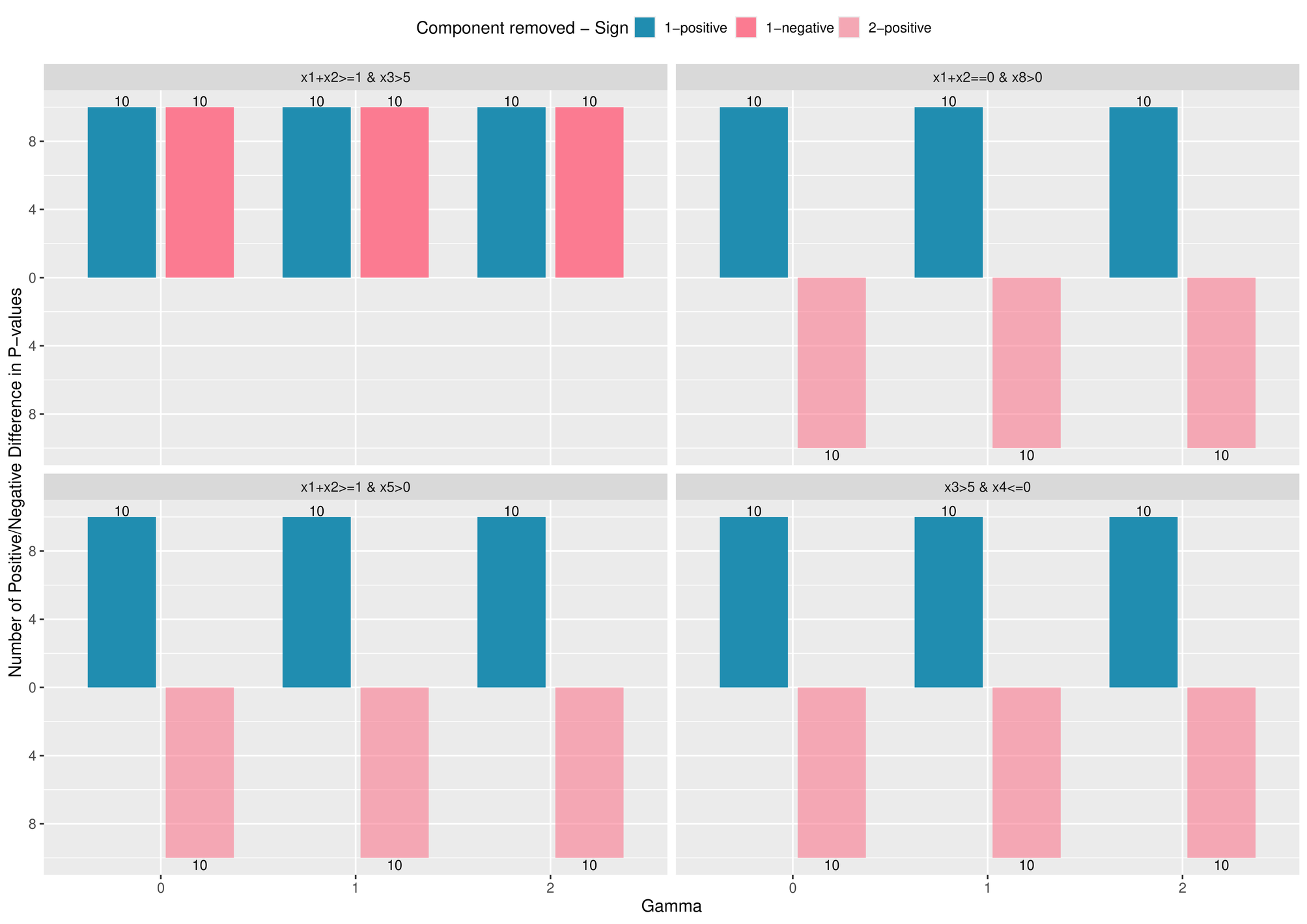}
    \caption{Results of rule decomposition analysis under different strengths of variable correlations ($\gamma=0,1,2$) for $\{N=10000,num.grp=5,k=30\}$.
    The pillars show the count of positive (above 0  vertical line) and negative (below 0 vertical line) $\Delta p$ corresponding to the first component removed (blue) and the second component (pink) removed. }
    \label{fig:p-5grp}
\end{figure}

\textbf{Results.} Figure \ref{fig:p-1grp},\ref{fig:p-3grp} and \ref{fig:p-5grp} show the number of positive and negative $\Delta p$ under different $\gamma$ for each specific type of rule and data setting.
For example, Figure \ref{fig:p-1grp} shows the results for the three types of rules of $\{N=2500,num.grp=1,k=10\}$:
\begin{itemize}
    \item  For the true rule $x_1=1$ and $x_2=1$ (upper left), we observe that when $\gamma=0$, removing $x_1$ (critical variable, blue pillars) results in a positive $\Delta p$ 7 times and a negative $\Delta p$ 3 times out of 10 randomized trials, demonstrating the importance of $x_1$. 
This importance does not decay as $\gamma$ increases, as we observe 10 times of positive $\Delta p$ when $\gamma=1$ and $\gamma=2$.
Similarly, removing $x_2$ (critical variable, pink pillars) yields 10, 9, and 9 times of positive $\Delta p$. 
Therefore, this true rule is more likely to survive in our rule analysis step. 

    \item  For the fake rule without correlation (upper right) $x_1=1  $ and $ x_8>0$, we find a clear contrast of removing $x_1$ against removing $x_8$. 
While $x_1 $ remains important to the complete rule (10 positive $\Delta p$), $x_8 $ exhibits irrelevance to the true subgroup with 10 negative $\Delta p$ across all $\gamma$. 
Hence, in all cases, this fake rule will be filtered out in our rule analysis step.

    \item  For the fake rule with correlation (bottom) $x_1=1  $ and $ x_5>0$, $x_1=0  $ and $ x_5\le0$, we find the contrast between critical variable and redundant variable remains, except for a few cases (1 and 3) of positive $\Delta p$ when the correlation is strong ($\gamma=2$), suggesting a small chance of falsely regarding $x_5$ as a critical variable and the rule survives in our analysis step. 
\end{itemize}

We find nearly identical results for the other two data settings (Figure \ref{fig:p-3grp} and \ref{fig:p-5grp}), where true rules all survive and other fake rules are all filtered out. 
Empirically, we can tell that although strong correlations may alter the positive $\Delta p$ signal of critical variables (lower left of Figure \ref{fig:p-1grp},  lower right of Figure \ref{fig:p-3grp}), the negative $\Delta p$ signal of redundant variables does not diminish dramatically as correlation increases. 
This is because when the critical variable is removed, the redundant variable can substitute for the removed variable to a certain degree, leading to the uncertainty of $\Delta p$.
However, removing the redundant variable always makes the revised rule closer to the true rule and hence a negative $\Delta p$. 
Even in rare cases where such rules survive, the strong correlation between variables in these rules can easily be identified by human experts and deemed invalid. 
For instance,  $x_1=0  $ and $ x_5\le0$ in the lower right of Figure \ref{fig:p-1grp}). 
Based on the above findings, we conclude that variable correlations do not substantially compromise the overall validity of our rule decomposition analysis.

\section{Descriptive statistics of ASD variables}
\label{apx:ASD_data}
\setcounter{figure}{0}  
\setcounter{table}{0}
\setlength{\tabcolsep}{3pt}

\begin{figure}[t]
	\centering
	\includegraphics[width=1\linewidth]{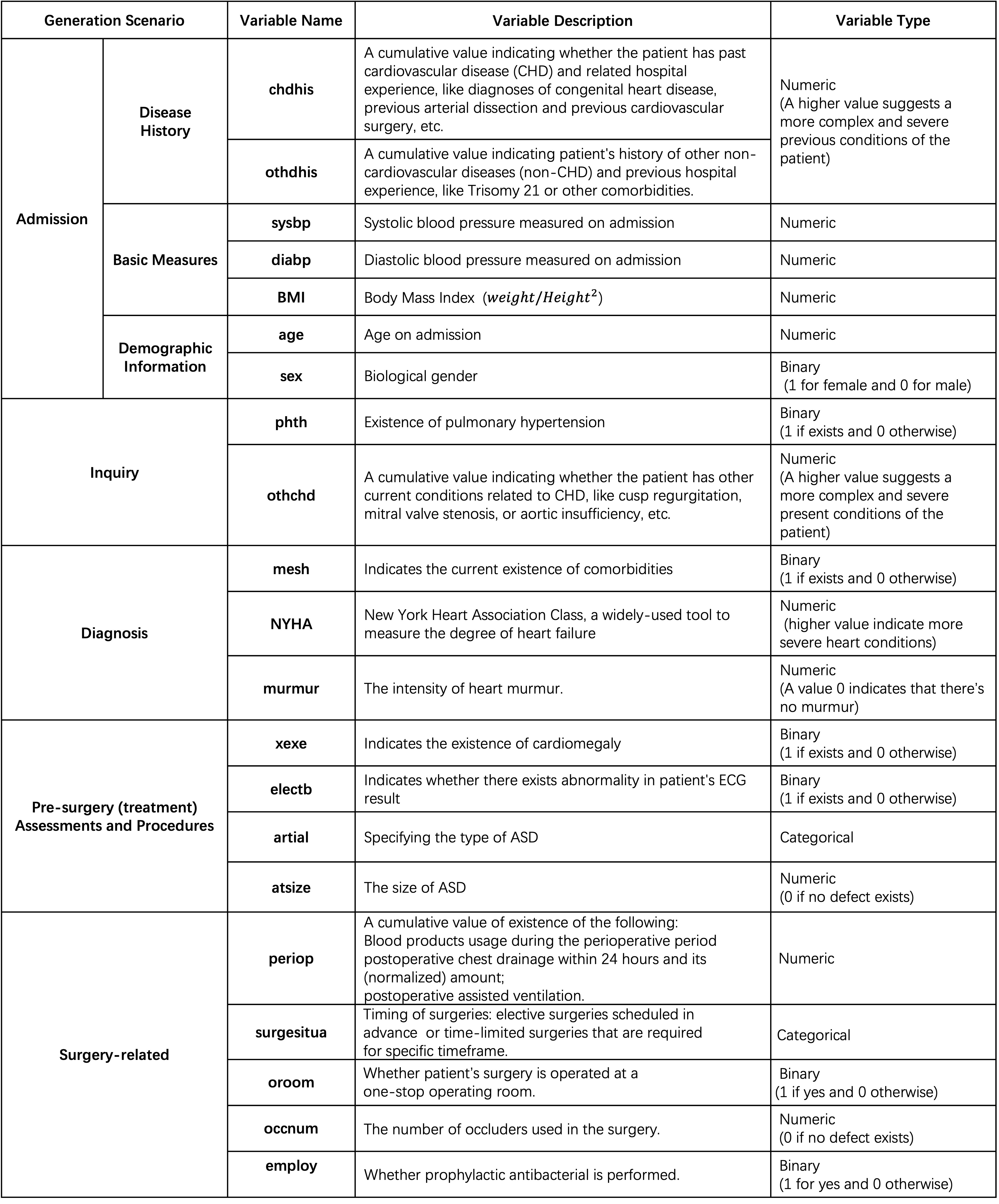}
 \caption{Description of 16 pre-treatment variables and 5 surgery-related variables.}
	\label{fig:realdata.des}	
\end{figure}

\begin{longtable}{lcrrrr}
\caption{Descriptive statistics of 16 pre-treatment variables. The last two columns show the \textit{p}-values from the two-sample \textit{t}-test and the standardized mean difference (SMD) between the PIC and MISC groups.}
\label{tab:des.stat.realdata1}\\
\hline 
\endfirsthead
\multicolumn{6}{c}{\itshape continued from previous page.} \\
\hline 
\endhead 
\hline 
\multicolumn{6}{c}{\itshape continued on next page}\\[2ex]
\endfoot
\hline 
\endlastfoot
            \textbf{Covariates} & \textbf{Level} & \textbf{PIC (positive)}&\textbf{MISC (negative)}& \textbf{$p$-value} & \textbf{SMD} \\ 
            & & N=2201& N=379& & \\ \hline
		\textbf{othdhis (\%)} &0& 2154 (97.9)&373(98.4)&0.757&0.077\\
		&1&41(1.9)&6(1.6)&\\
		&2&5(0.2)&0(0.0)&\\
		&3&1(0.0)&0(0.0)&\\
		\textbf{chdhis(\%)}&0&13(0.6)&1(0.3)&0.498&0.106\\
		&1&1032(46.9)&195(51.5)&\\
		&2&1147(52.1)&182(48.0)&\\
		&3&8(0.4)&1(0.3)  &  \\
		 & 4 & 1 (0.0) & 0 (0.0)&  \\
		\textbf{othchd (\%)} & 0 & 787 (35.8) & 147 (38.8) & 0.002 & 0.221 \\
		 & 1 & 4 (0.2) & 2 (0.5)&  \\
		 & 2 & 25 (1.1)& 2 (0.5)&  \\
		 & 3 & 2 (0.1) & 0 (0.0)&  \\
		 & 4 & 13 (0.6)& 9 (2.4)&  \\
		 & 5 & 1311 (59.6) & 210 (55.4)&  \\
		 & 6 & 57 (2.6)& 7 (1.8)&  \\
		 & 7 & 1 (0.0) & 2 (0.5)&  \\
		 & 8 & 1 (0.0) & 0 (0.0)&  \\
		\textbf{atrial (\%)} & 0 & 45 (2.0)& 4 (1.1) & \multicolumn{1}{c}{0.011} & 0.221 \\
		 & 1 & 241 (11.0) & 33 (8.7)  &  \\
		 & 2 & 1861 (84.6) & 340 (89.9)&  \\
		 & 3 & 52 (2.4)& 1 (0.3)&  \\
      \textbf{NYHA (\%)} & 1& 636 (29.0) & 261 (68.9) & $<$0.001 & 0.92 \\
		 & 2& 1491 (67.9) & 100 (26.4)&\\
		 & 3& 68 (3.1) & 18 (4.7)&\\
		\textbf{mesh (\%)} & 0& 2071 (94.1) & 318 (83.9) & $<$0.001 & 0.33 \\
		 & 1& 130 (5.9)& 61 (16.1) &\\
		\textbf{phth (\%)} & 0& 2092 (95.0) & 310 (81.8) & $<$0.001 & 0.423 \\
		 & 1& 109 (5.0)& 69 (18.2) &\\
		\textbf{murmur (\%)} & 0& 568 (25.8) & 145 (38.3) & $<$0.001 & 0.341 \\
		 & 1& 33 (1.5) & 11 (2.9)&\\
		 & 2& 302 (13.7) & 61 (16.1) &\\
		 & 3& 1298 (59.0) & 162 (42.7)&\\
		\textbf{electb (\%)} & 0& 1401 (63.7) & 108 (28.5) & $<$0.001 & 0.754 \\
		 & 1& 800 (36.3) & 271 (71.5)&\\
		\textbf{xece (\%)} & 0& 1130 (51.3) & 311 (82.1) & $<$0.001 & 0.689 \\
        & 1& 1071 (48.7)& 68 (17.9)& &\\
        \textbf{sex (\%)} & 0& 985 (44.8) & 148 (39.1) & 0.044& 0.116 \\
        & 1& 1216 (55.2) & 231 (60.9)& &\\
        \textbf{atsize0 (mean (SD))} && 9.41 (6.39) & 17.31 (8.61) & $<$0.001  & 1.042 \\
        \textbf{sysbp (mean (SD))} & & 203.04 (13.33) & 99.58 (12.07) & $<$0.001 & 0.272 \\
        \textbf{diabp (mean (SD))} & & 63.75 (10.32) & 62.05 (10.68) & 0.004 & 0.162 \\
        \textbf{age (mean (SD))} & & 10.26 (9.77) & 8.74 (10.26) & 0.006 & 0.152 \\
        \textbf{BMI (mean (SD))} & & 17.54 (5.1) & 16.44 (3.60) & 0.001 & 0.248 \\
	\hline
\end{longtable}

\begin{table}[h]
    \centering
    \caption{Descriptive statistics of five surgery-related variables and outcome variable \textit{HFD1Y}.
    }
    
    \begin{tabular}{ccccc}
    \hline
            Vairables &  \textbf{PIC (positive)}&  \textbf{MISC (negative)}& \textbf{$p$-value}  & \textbf{SMD}\\
  (mean (SD))& N=2201& N=379& &\\
    \hline
         \textit{surgesitua} & 1.00 (0.06) & 1.01 (0.09) & 0.171 & 0.064 \\
		oroom& 0.22 (0.42) & 0.31 (0.46) & $<$0.001     & 0.207 \\
		\textit{occnum}  & 1.00 (0.13) & 0.10 (0.30) & $<$0.001     & 3.858 \\
		\textit{periop}  & 1.12 (0.37) & 3.09 (0.82) & $<$0.001     & 3.111 \\
            \textit{employ} &  0.64 (0.48) &  0.70 (0.46) & 0.016& 0.136\\
            HFD1Y   & 360.05 (4.75) & 352.49 (6.17) & $<$0.001     & 1.373 \\
            \hline
    \end{tabular}
    \label{tab:des.stat.realdata2}
\end{table}
\section{Propensity score matching of the ASD data}\label{apx:psm}
\setcounter{figure}{0}  
\setcounter{table}{0}

\begin{figure}[h]
	\centering
	\includegraphics[scale = 0.6]{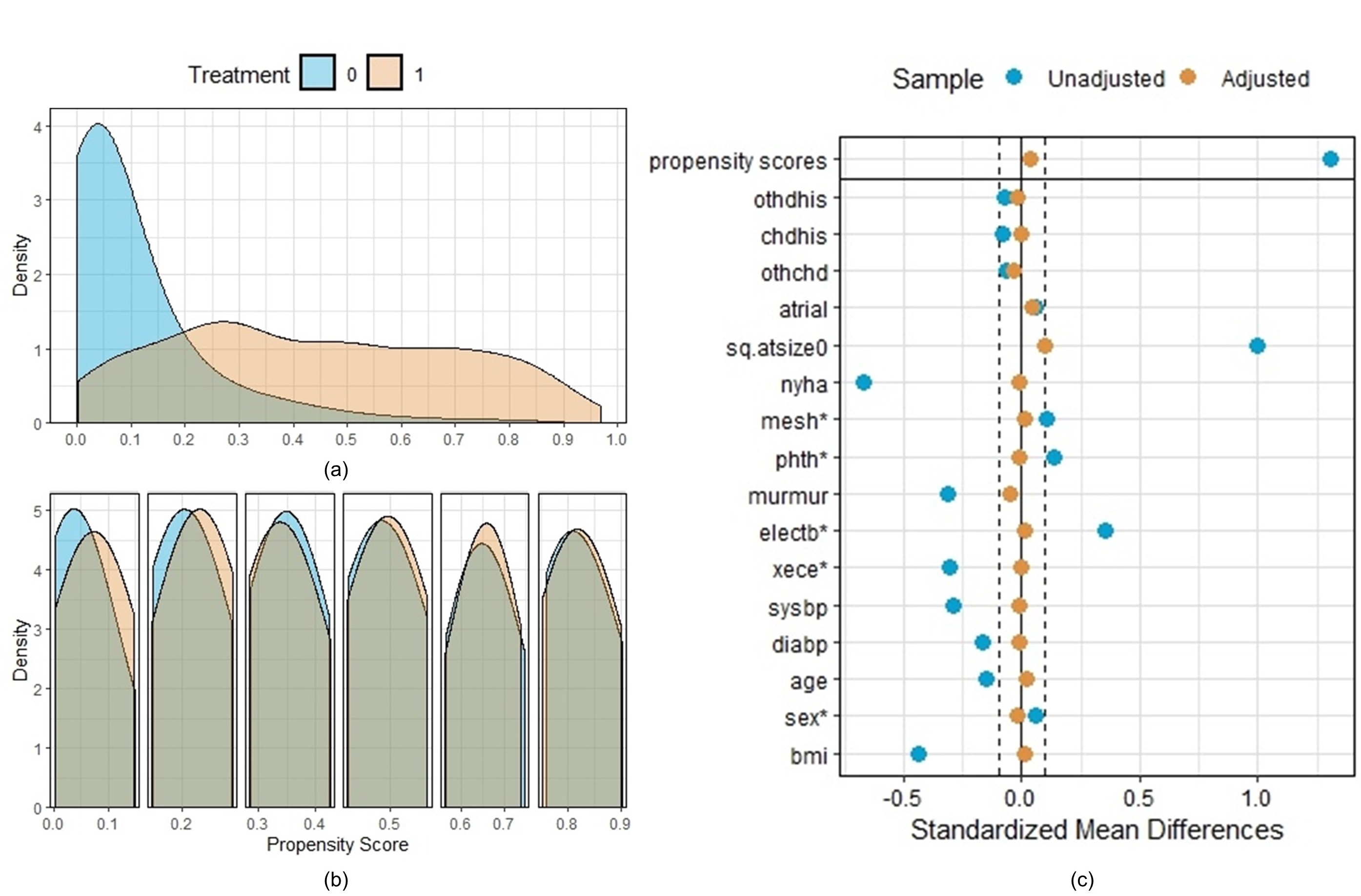}	
	\caption{Details of propensity score matching on the ASD data, including the density of propensity scores before (a) and after (b) matching and the love plot (c) of both treatment groups.}
	\label{fig:psm}
\end{figure} 
We use the R package \textit{MatchIt} \cite{ho2011package} to perform the propensity score matching procedure on the pre-treatment variables of the ASD data. 
Note that we recode the negative treatment to $0$ instead of $-1$. 
The procedure involves three core aspects: propensity score estimation, data matching, and balance diagnostics. 
We measure the distance between samples in terms of propensity scores and try various combinations of propensity score estimation models and matching methods. 
Our final choice goes to a covariate balancing propensity score \cite{imai2014covariate} model with the following implementation formula to estimate propensity scores, by which the samples are stratified into subclasses for matching:  

\textit{treatment = othdhis + chdhis + othchd + atrial + $\sqrt{atsize0}$ + NYHA +  mesh + phth + murmur + electb + xece + sysbp + diabp + age +  sex + BMI + employ.}

This combination is able to achieve

i) good overlap within each subclass, see Figure \ref{fig:psm}(b). 
The improvement of overlap before (Figure \ref{fig:psm}(a)) and after (Figure \ref{fig:psm}(b)) the matching is obvious;
ii) balanced covariates between the two treatment groups. Before the matching, there are only five variables that are considered balanced between the two groups, having a standardized mean difference (SMD) within 10\%. 
After the matching, all covariates are balanced other than the square root of atsize0 (sq.atsize0), whose SMD value is slightly above the threshold (0.1005). 
Here, we consider this difference to be acceptable as it is what we can best achieve from all attempts. 
See the love plot in Figure \ref{fig:psm}(c) for the comparison of covariate balance between treatments before and after the matching;
iii) the fewest observations discarded. 
The matching process retains a total of 2554 observations, with 2180 for positive treatment (PIC) and 374 for negative treatment (MISC). 
We then apply CRL on this matched data set.

\end{document}